\documentclass[journal, twoside, twocolumn]{IEEEtran}
\usepackage{amsmath,amsfonts}
\usepackage{algorithmic}
\usepackage{algorithm}
\usepackage{array}
\usepackage[caption=false,font=normalsize,labelfont=sf,textfont=sf]{subfig}
\usepackage{textcomp}
\usepackage{multirow}
\usepackage{booktabs}
\usepackage{stfloats}
\usepackage{url}
\usepackage{balance}
\usepackage{verbatim}
\usepackage{graphicx}
\usepackage{setspace}
\usepackage{cite}
\usepackage{threeparttable}
\usepackage{color}
\usepackage{tcolorbox}
\newtheorem{Def}{Definition}

\begin{document}

\title{Self-Supervised Temporal Graph Learning with Temporal and Structural Intensity Alignment}

\author{Meng~Liu,~Ke~Liang,~Yawei~Zhao,~Wenxuan~Tu,~Sihang~Zhou,\\~Xinbiao~Gan,~Xinwang~Liu$^{\ast}$,~\IEEEmembership{Senior~Member,~IEEE},~Kunlun~He$^{\ast}$
    
    \thanks{$^{\ast}$ Corresponding author.}
    \thanks{Meng Liu, Ke Liang, Wenxuan Tu, Xinbiao Gan, and Xinwang Liu are with the School of Computer, National University of Defense Technology, Changsha, China. E-mail: {mengliuedu@163.com, xinwangliu@nudt.edu.cn}.}
    \thanks{Yawei Zhao, and Kunlun He are with Medical Big Data Research Center, Chinese PLA General Hospital, Beijing, China. E-mail: {csyawei.zhao@gmail.com, kunlunhe@plagh.org}.}
    \thanks{Sihang Zhou is with the College of Intelligence Science and Technology, National University of Defense Technology, Changsha, China.}
    \thanks{This work has been submitted to the IEEE for possible publication. Copyright may be transferred without notice, after which this version may no longer be accessible.}
}
\markboth{IEEE Transactions on Neural Networks and Learning Systems}%
{IEEE Transactions on Neural Networks and Learning Systems}

\maketitle

\begin{abstract}
Temporal graph learning aims to generate high-quality representations for graph-based tasks with dynamic information, which has recently garnered increasing attention. In contrast to static graphs, temporal graph are typically organized as node interaction sequences over continuous time rather than an adjacency matrix. Most temporal graph learning methods model current interactions by incorporating historical neighborhood. However, such methods only consider first-order temporal information while disregarding crucial high-order structural information, resulting in sub-optimal performance. To address this issue, we propose a self-supervised method called S2T for temporal graph learning, which extracts both temporal and structural information to learn more informative node representations. Notably, the initial node representations combine first-order temporal and high-order structural information differently to calculate two conditional intensities. An alignment loss is then introduced to optimize the node representations, narrowing the gap between the two intensities and making them more informative. Concretely, in addition to modeling temporal information using historical neighbor sequences, we further consider structural knowledge at both local and global levels. At the local level, we generate structural intensity by aggregating features from high-order neighbor sequences. At the global level, a global representation is generated based on all nodes to adjust the structural intensity according to the active statuses on different nodes. Extensive experiments demonstrate that the proposed model S2T achieves at most 10.13\% performance improvement compared with the state-of-the-art competitors on several datasets.
\end{abstract}

\begin{IEEEkeywords}
Temporal graph learning, self-supervised learning, conditional intensity alignment.
\end{IEEEkeywords}

\section{Introduction}
\IEEEPARstart{G}{raph} learning has drawn increasing attention in recent years \cite{cui2018survey, wu2020comprehensive} due to the prevalence of graph-based representations in various real-world scenarios, such as web graphs, social graphs, and citation graphs \cite{qiu2018network, wang2019neural, LiangTNNLS}. Traditional graph learning methods are based on static graphs, and they usually use adjacency matrices to generate node representations by aggregating neighborhood features \cite{ou2016asymmetric, li2021learning}.

In contrast to static graphs, temporal graphs are structured based on node interaction sequences over continuous time, rather than relying on adjacency matrices. In many real-world cases, graphs exhibit node interactions accompanied by timestamps. It becomes crucial for models to capture and incorporate temporal information into node representations to facilitate accurate prediction of future interactions.

It is worth noting that temporal graph-based methods can hardly generate node representations directly using the adjacency matrix to aggregate neighbor information like a static graph-based method \cite{zhang2021we, bo2020structural}. Due to the different data form of the temporal graph that sorts node interactions by time, temporal methods are trained in batches of data \cite{huang2022discovering}. Consequently, these methods typically store neighbors in interaction sequences and model future interactions from historical information. However, such methods merely consider the first-order temporal information while ignoring the important high-order structural neighborhood, leading to sub-optimal performance.

To solve this issue, by extracting both \textbf{T}emporal and structural information to learn more informative node representations, we propose a \textbf{S}elf-\textbf{S}upervised method termed S2T for \textbf{T}emporal graph learning. Note that the first-order temporal information and the high-order structural neighborhood are combined in different ways by the initial node representations to calculate two conditional intensities, respectively. Then the alignment loss is introduced to optimize the node representations to be more informative by narrowing the gap between the two intensities. More specially, for temporal information modeling, we leverage the Hawkes process \cite{hawkes1971point} to calculate the temporal intensity between two nodes. Besides considering the two nodes' features, the Hawkes process also considers the effect of their historical neighbors on their future interactions.

On the other hand, we further extract the structural information, which can be divided into the local and global level. When capturing the local structural information, we first utilize GNN to generate node representations by aggregating the high-order neighborhood features. After that, the global structural information is extracted to enhance long-tail nodes. In particular, a global representation generated based on all nodes is proposed, which is used to update node representations according to active statuses on different nodes. After the structural intensity is calculated based on node representations, we also construct a global parameter to assign importance weights for different dimensions of the structural intensity vector. Finally, in addition to the task loss, we utilize the alignment loss to narrow the gap between the temporal and structural intensity vectors and impose constraints on global representation and parameters, which constitute the total loss function.

We conduct extensive experiments to compare our method S2T with the state-of-the-art competitors on several datasets, the results demonstrate that S2T achieves at most 10.13\% performance improvement. Furthermore, the ablation and parameter analysis shows the effectiveness of our model.

In conclusion, the contributions are summarized as follows:

(1) We extract both temporal and structural information to obtain different conditional intensities and introduce the alignment loss to narrow their gap for learning more informative node representations.

(2) To enhance the information on the long-tail nodes, we capture the global structural information as an augmentation of the local structural module.

(3) We compare S2T with multiple methods in several datasets and demonstrate the performance of our method.

\section{Related Work}
\label{related work}

\begin{figure}[t]
    \centering
    \includegraphics[width=0.48\textwidth]{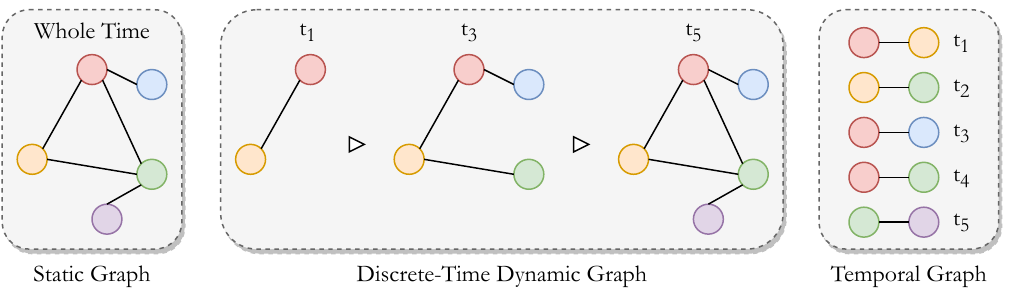}
    \caption{Static graphs, discrete dynamic graphs, and temporal graphs. Static graphs have only one whole time, which can be considered as the final moment. Discrete dynamic graphs intercept the current state of the graph at equal intervals to generate a static snapshot of the corresponding moment. Temporal graphs record the time of each node interaction, which is continuous in a realistic sense, and are also known as continuous-time dynamic graphs.}
    \label{different graph}
\end{figure}

\subsection{Graph Learning}

Graph Learning is an important technology which can be used for many fields, such as interest recommendation \cite{wu2022gcrec}, biological informatics \cite{hu2023scdfc}, knowledge graph \cite{liang2023learn, ji2021survey, meng2023sarf}, smart city \cite{gao2023spatiotemporal}, and community detection \cite{CCGC}, etc. In these fields, researchers denote people or items as nodes, and the relationships between them are considered as edges. In this way, many real-world relationships can be represented as graphs. Deep graph learning aims to mine the important information or laws in these graphs, which generates representation vector for each node. Such representations can be utilized for many downstream tasks, such as link prediction, node classification, node clustering, etc. These downstream tasks can be seen as the application pattern of the above fields. Based on this, graph learning can also help multi-modal modeling \cite{liang2022reasoning, liang2023structure}, information fusion \cite{DMG_ICML, zhou2023multi}, and relation discovery \cite{Liangke_SymCLKG_TKDE, mo2023multiplex}, etc. Mechanistic issues about graph learning such as security, trustworthy, and interpretability are also gradually coming to the attention of researchers \cite{yu2023g}.

Further, graph data has many different classifications, such as heterogeneous graphs \cite{wang2019heterogeneous}, hypergraphs \cite{wu2023hypergraph, wu2023simplicial}, and so on. Here, we mainly discuss the classification based on temporal information, i.e., static and dynamic graphs. The most essential difference between them is whether the data contains interaction time information \cite{fang2022scalable, gan2022multigraph}.

\subsection{Static and Temporal Graphs}

As shown in Figure \ref{different graph}, we discuss the different types of graph data here. Traditional graph learning methods learn node representations on \textbf{static graphs}, which focus on the graph topology or adjacency matrix \cite{song2022graph, Wan_Liu_Liu_Wang_Wen_Liang_Zhu_Liu_Zhou_2023}. In these graphs, nodes and edges will not change, and there is no concept of time \cite{lin2022prototypical, CONVERT}.

To name a few, DeepWalk \cite{perozzi2014deepwalk} performs random walks over the graph to learn node embeddings (also called representations). node2vec \cite{grover2016node2vec} conducts random walks on the graph using breadth-first and depth-first strategies to balance neighborhood information of different orders. VGAE \cite{kipf2016variational} migrates variational auto-encoders to graph data and use encoder-decoder module to reconstruct graph information. GraphSAGE \cite{hamilton2017inductive} leverages an aggregation function to sample and combine features from a node's local neighborhood. PGExplainer \cite{luo2020parameterized} adopts a deep neural network to parameterize the generation process of explanations, which can explain multiple instances collectively. InfoGCL \cite{xu2021infogcl} discusses how graph information is transformed and transferred during the contrastive learning process. NeuralSparse \cite{zheng2021node} is a supervised graph sparsification technique that improves generalization power by learning to remove potentially task-irrelevant edges from input graphs.

In addition, many real-world data contain dynamic interactions, thus graph learning methods based on \textbf{dynamic graphs} are becoming popular \cite{cui2022dygcn, TGC_ML}. Concretely, dynamic graphs can also be divided into discrete graphs (also called discrete-time dynamic graphs, DTDG) and temporal graphs (also called continuous-time dynamic graphs, CTDG). 

A \textbf{discrete graph} usually contains multiple static snapshots, each snapshot is a slice of the whole graph at a certain timestamp, which can be regarded as a static graph. When multiple snapshots are combined, there is a time sequence among them. Because each static snapshot is computationally equivalent to that of a static graph, which makes the computation significantly less efficient, only a small amount of work is performed on discrete graphs. Such as EvolveGCN \cite{pareja2020evolvegcn} uses the RNN model to update the parameters of GCN for future snapshots. DySAT \cite{sankar2020dysat} combines graph structure and dynamic information to generate self-weighted node representations.

Unlike discrete graphs, a \textbf{temporal graph} no longer observes graph evolution over time from a macro perspective but focuses on each node interaction. For example, CTDNE \cite{nguyen2018continuous} performs a random walk on graphs to model temporal ordered sequences of node walks. HTNE \cite{zuo2018embedding} is the first to utilize the Hawkes process to model historical events on temporal graphs. MMDNE \cite{lu2019temporal} models graph evolution over time from both macro and micro perspectives. STAR \cite{xu2019spatio} extracts the vector representation of neighborhood by sampling and aggregating local neighbor nodes. AdaNN \cite{xu2019adaptive} learns node attribute information by combining the node and its neighbors, and extracts network topology information with a random walk strategy. TGAT \cite{xu2020inductive} replaces traditional modeling form of self-attention with interaction temporal encoding. MNCI \cite{liu2021inductive} mines community and neighborhood influences to generate node representations inductively. TSNet \cite{zheng2021node} jointly learns temporal and structural features for node classification from the sparsified temporal graphs. TRRN \cite{xu2021transformer} is a transformer-style relational reasoning network with dynamic memory updating. TREND \cite{wen2022trend} replaces the Hawkes process with GNN to model the temporal information. DynG2G \cite{xu2022dyng2g} applies an inductive feedforward encoder trained with node triplet energy-based ranking loss. STGAN \cite{deng2022graph} is a spatio-temporal generator to predict the normal traffic dynamics and a spatio-temporal discriminator to determine whether an input sequence is real or not. DyGCN \cite{cui2022dygcn} is naturally generalized to a dynamic setting in an efficient manner, which propagates the change in topological structure and neighborhood embeddings along the graph to update the node embeddings. DyCPM \cite{duan2023dynamic} not only generates low-dimensional embedding vectors of nodes, but also aggregates the structural information and temporal information of two kinds of edges. TGC \cite{TGC_ML} first discusses the deep temporal graph clustering task. TMac \cite{liu2023tmac} introduces the temporal multi-modal graph network for acoustic event classification. DyExplainer \cite{wang2023dyexplainer} is a novel approach to explaining dynamic GNNs on the fly.

\subsection{Differences with Existing Methods}

Although the methods mentioned above have demonstrated their effectiveness, they primarily consider either one-order temporal information or high-order structural neighborhood. However, there is room for further improvement. Integrating both types of information remains an open problem, prompting us to propose the S2T method as a solution.

Moreover, we posit that the proposed S2T method introduces a novel perspective to address the current challenges in temporal graph learning. Acquiring higher-order structural information has traditionally proven challenging for temporal graph methods. For instance, approaches like HTNE, JODIE, and MNCI only account for first-order neighbors, while TREND is constrained by the neighborhood orders, resulting in substantial computational effort when mining higher-order neighbor information. Thus, we strive to strike a balance between these two aspects.

To capture low-order neighbor structures such as second-order and third-order, the computational load remains manageable, allowing us to employ GNNs for information propagation. However, for higher-order structural neighborhood, the aforementioned approach becomes impractical. As a result, we introduce global variables to preserve such information to the greatest extent possible.

By adopting this approach, we maximize the enhancement of structural information within the temporal graph model while keeping the training complexity reasonably low. This contribution forms the core of our research endeavor.

\section{Method}
\label{method}

In this part, we first introduce the overall framework of S2T and then denote some preliminaries. After that, we describe each component module in detail.

\subsection{Overall Framework}

\begin{figure*}[htbp]
    \centering
    \includegraphics[width=0.95\textwidth]{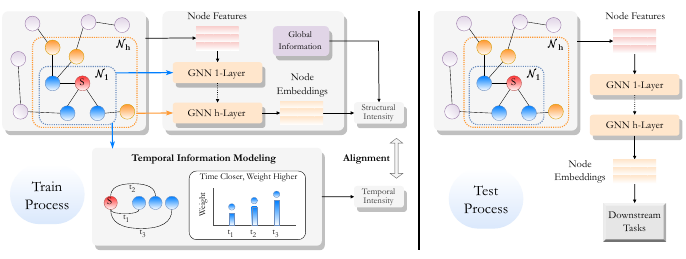}
    \caption{Overall Framework of S2T. (1) During training, we utilize multi-layer GNNs to generate node embeddings and incorporate global information for computing structural intensity. The Hawkes process modeling first-order time information is also introduced to compute the temporal intensity. The parameters of GNNs are optimized by constraining the alignment of two intensities. (2) During testing, since the parameters of GNNs have been optimized, the model feeds node features directly into the GNNs to generate node embeddings for downstream tasks.}
    \label{intro}
\end{figure*}

As shown in Figure \ref{intro}, our method S2T can be divided into several main parts: temporal information modeling, structural information modeling, and loss function, where a local module and a global module work together to realize structural information modeling. Then the alignment loss is introduced into the loss function, which aligns temporal and structure information.

\subsection{Preliminaries}

First, we define the temporal graph based on the timestamps accompanying the node interactions.

\begin{Def}
    \textbf{Temporal Graph.}
    Given a temporal graph $G=(V, E, T, X)$, where $V$ and $E$ denote the set of nodes and edges (called interactions here), $T$ denotes the timestamp set of node interactions, and $X$ denotes the node features. If an edge exists between node $x$ and $y$, this means that they have interacted at least once, i.e.,  $T_{x, y} = \{(x, y, t_1), (x, y, t_2), \cdots, (x, y, t_n)\}$.
\end{Def}

When two nodes interact, we call them neighbors. Note that in temporal graphs, the concept of interaction replaces the concept of edges, and multiple interactions can occur between two nodes.

\begin{Def}
    \textbf{Historical Neighbor Sequence.}
    For each node $x$, there will be a historical neighbor sequence $N_x$, which stores all interactions of $x$, i.e., $N_x = \{ (y_1,t_1), (y_2,t_2), \cdots, (y_n, t_n) \}$.
\end{Def}

In a temporal graph, one interaction data is stored as a tuple of $(x,y,t)$, which means that the two nodes $x$ and $y$ interact at time $t$. In the actual training, we feed these interaction data into the model in batches. Our objective is to conduct a mapping function $F$ that converts high-dimensional sparse graph-structured data $G$ into low-dimensional dense node representations $Z$.

\subsection{Temporal Information Modeling}

To maintain the paper's continuity, we first introduce the temporal module and then introduce the structural module.

Given two nodes $x$ and $y$, we can indicate the likelihood of their interaction by calculating the conditional intensity between them. There are two ways to obtain it, here we discuss the first way: modeling temporal information with the Hawkes process \cite{hawkes1971point}. Such a point process considers a node's historical neighbors will influence the node's future interactions, and this influence decays over time.

Define $z_x$ and $z_y$ to denote the representations of node $x$ and $y$ respectively, which is obtained by a simple linear mapping of their features. Their temporal interaction intensity $\lambda^T_{(x,y)}(t)$ can be calculated as follows,
\begin{equation}
    \lambda^T_{(x,y)}(t) = \mu_{(x, y)} + \sum_{i \in N_x}\alpha_{(i, y)} \mu_{(i, y)}  + \sum_{i \in N_y}\alpha_{(i, x)} \mu_{(i, x)},
    \label{hawkes lambda}
\end{equation}
\begin{equation}
    \mu_{(x, y)} = - || z_x - z_y ||^2, \quad \alpha_{(i, y)} = s_{(i,x)} \cdot f(t_c-t_i).
\end{equation}

This intensity can be divided into two parts: (1) the first part is the base intensity between two nodes without any external influence, i.e., $\mu_{(x, y)}$; (2) the second part is the Hawkes intensity that focuses on how a node's neighbors influence another node, where $i$ denotes the neighbor in the sequence, i.e., $\sum_{i \in N_x}\alpha_{(i, y)} \mu_{(i, y)}$.

In the Hawkes intensity, $\alpha_{(i, y)}$ measures the influence of a single neighbor node $i$ of $x$ on $y$, and this influence is weighted by two aspects. On the one hand, $s_{(i,x)}$ is the similarity weight between neighbor $i$ and source node $x$ in the neighbor sequence $N_x$, i.e., $s_{(i,x)} = \frac{\exp(\mu_{(i, x)})}{\sum_{i' \in N_x}\exp(\mu_{(i', x)})}$. This similarity weight means that although we calculate the influence of each neighbor $i$ in $N_x$ on $y$, we also need to consider the corresponding weights $s_{(i,x)}$ for different $i$ in $N_x$. Note that in the Hawkes intensity, both $\mu_{(i, y)}$ and $\mu_{(i, x)}$ appear, and their roles are different. On the other hand, the function $f(t_c-t_i)$ considers the interaction time interval between $i$ and $x$, i.e., $f(t_c-t_i) = \exp(-\delta_t(t_c-t_i))$, where $\delta_t$ is a learnable parameter. In this function, neighbors that interact closer to the current time $t_c$ are given more weight.

In addition, the total number of neighbors may vary from node to node. In actual training, if we obtain all of its neighbors for each node, the computational pattern of each batch can not be fixed, which brings great computational inconvenience. Referencing previous works \cite{zuo2018embedding, hu2020graph, liu2022embtemporal, wen2022trend} and our experiments, we fix the sequence length $S$ of node neighbors and select the latest $S$ neighbors for each node at each timestamp instead of full neighbors.

\subsection{Local Structural Information Modeling}

In addition to the Hawkes process, the GNN model can also be used to calculate conditional intensity. Unlike the Hawkes process which focuses on the temporal information of the first-order neighbors, GNN is more concerned with aggregating information about the high-order neighbors. For each node $x$ at time $t$, we construct $l$ GNN layers to generate its representation $z_x^{t,l}$ as follows,
\begin{equation}
    z_x^{t,l} = \sigma(z_x^{(t,l-1)}W_{S}^l + \sum_{i \in N_x}z_i^{(t_i, l-1)}W_{N}^l \odot k(t_c-t_i)),
    \label{GNN}
\end{equation}
\begin{equation}
    k(t_c-t_i) = \frac{t_c-t_i}{\sum_{i' \in N_x}t_c-t_{i'}},
\end{equation}
where $W_S^l$ and $W_N^l$ are learnable parameters, $\odot$ denotes element-wise multiplication, $\sigma$ is the sigmoid function, and $k(t_c-t_i)$ is used to generate normalized weights for the interaction time intervals of different neighbors. The first layer's input $h_x^{t,0}$ is a simple linear mapping of node features, and the final layer's output $h_x^{t,l}$ is the aggregated representation containing the $l$-order neighborhood information. Note that both Hawkes intensity and GNN intensity utilize the original linear mapping representations as input, and we utilize generate representations based on GNN as the final output.

Given two nodes, their local conditional intensity measures how similar the information they aggregated is, which can be calculated as follows,
\begin{equation}
    \lambda^S_{(x,y)}(t) = -||z_x^t - z_y^t||^2 \odot \omega_g,
    \label{local intensity}
\end{equation}
where $\omega_g$ is a global parameter that will be described below. For brevity, we denote the final layer's output $z_x^{t,l}$ as $z_x^t$ and omit $l$.

\subsection{Global Structural Information Modeling}
\label{global modeling}

After calculating the node representations based on GNN, we construct a global module to enhance the structural information modeling. Firstly, let us discuss why we need information enhancement.

In most graphs, there are always a large number of long-tail nodes that interact infrequently but are the most common in the graph. Due to their limited interactions, it is hard to find sufficient data to generate their representations. Previous works usually aggregate high-order neighbor information to enhance their representations, which is consistent with the purpose of our GNN module above. But in addition, we worry that too much external high-order information will dominate instead, bringing unnecessary noise to long-tail nodes. Therefore, we generate a global representation that provides partial basic information for these nodes.

\subsubsection{Global Representation}

Global representation, as an abbreviated expression for the whole graph environment, is updated based on all nodes. In the graph, only a small number of nodes are high-active due to their large number of interactions that influence the graph evolution, while most long-tail nodes are vulnerable to the whole graph environment. Using global representation to fill basic information for long-tail nodes can ensure their representations are more suitable for unsupervised scenarios.

Here we introduce the concept of node active status from the LT model \cite{granovetter1978threshold} in the information propagation field \cite{li2018infmax, tian2011new}. A node's active status varies with its interaction frequency and can be used to measure how active a node is in the global environment. To be specific, the node active status can be used in two parts: (1) control the update of the global representation; (2) control the weight of global representation providing information to nodes.

For the first part, the global representation $z_g$ doesn't contain any information when it is initialized, and it needs to be updated by nodes. Note that in a temporal graph, nodes are trained in batches according to the interaction order. When a batch of nodes is fed into the model, the global representation can be updated as follows,
\begin{equation}
    z_g := z_g + g_g^t \odot z_x^t, \quad g_g^t = \theta_d \cdot |N_x^t|.
    \label{global update}
\end{equation}

In this equation, $\theta_d$ is a learnable parameter. $|N_x^t|$ is the number of neighbors that node $x$ interacts with at time $t$ and we call it node dynamics here. $g_g^t$ determines how much $x$ updates the global representation. In general, the more active a node $x$ is, the more influence it has on the global environment, so its corresponding weight $g_g^t$ is larger.

For the second part, the global representation is generated to enhance the long-tail nodes, thus it needs to add to the node representations. In contrast, the less active a node is, the more basic information it needs from the global representation. As for nodes with high active status, they have a lot of interactive information and do not need much data enhancement. Thus the update of node representations can be formed as follows,
\begin{equation}
    z_x^t := z_x^t + g_x^t \odot z_g, \quad g_x^t = \theta_d / |N_x^t|.
    \label{node emb update}
\end{equation}

Compared to Eq. \ref{global update}, the same parameter $\theta_d$ is used in Eq. \ref{node emb update}, while the node dynamics are set to the reciprocal, i.e., $1/|N_x^t|$. In one batch training, we first update the global representation, and then update the representation. The specific training flow is shown in Algorithm. \ref{S2T code}.

By enhancing the long tail nodes, the model can learn more reliable node representations. The more long-tail nodes in a graph, the more obvious the effect is, which is demonstrated in the following experiment subsection \ref{linkpre}.

\subsubsection{Global Parameter}
As mentioned in Eq. (\ref{local intensity}), after calculating the local intensity, we also construct a global parameter $\omega_g$ to assign importance weights for different dimensions of the intensity vector $\lambda^S_{x,y}(t)$. More specially, this global parameter can fine adjust the different dimensions of conditional intensity through a set of scaling and shifting operations so that those dimensions that are more likely to reflect node interaction are amplified. We first construct a simple learnable parameter $\theta_l$ , and then leverage the Feature-wise Linear Modulation (FiLM) layer \cite{perez2018film} to construct $\omega_g$,
\begin{equation}
    \omega_g = (\alpha^{(x,y,t)} + 1) \cdot \theta_l + \beta^{(x,y,t)},
    \label{omega g}
\end{equation}
\begin{equation}
    \alpha^{(x,y,t)} = \sigma(W_{\alpha} \cdot (z_x^t || z_y^t) + b_{\alpha}),
\end{equation}
\begin{equation}
    \beta^{(x,y,t)} = \sigma(W_{\beta} \cdot (z_x^t || z_y^t) + b_{\beta}).
\end{equation}

FiLM layer consists two parts: a scaling operator $\alpha^{(x,y,t)}$ and a shifting operator $\beta^{(x,y,t)}$, where $W_{\alpha}$, $b_{\alpha}$, $W_{\beta}$, $b_{\beta}$ are learnable parameters. Note that parameter enhancement by combining node representations allows the new parameters $\omega_g$ to better understand the meaning of each dimension of node representations. In this way, the parameter can finely adjust the importance of the different dimensions of the intensity vector, so that the conditional intensity can better reflect the possibility of node interaction.

\subsection{Loss Function}

\subsubsection{Task Loss}

The local conditional intensity calculated above is used to construct the classic loss function for link prediction \cite{chen2018pme} and we utilize the GNN-based node representations as final input into the loss function $L_{task}$ as follows,
\begin{equation}
    L_{task} = -\log\sigma(\lambda^S_{(x, y)}(t)) - \sum_{k \sim P_x} \log \sigma (1-\lambda^S_{(x,k)}(t)).
    \label{lt1}
\end{equation}

In this loss function, we introduce the negative sampling technology \cite{mikolov2013distributed} to generate the positive pair and negative pair. The positive pair contains nodes $x$ and $y$, their local intensity can be used to measure how likely they are to interact. For the negative pair, we introduce the negative sampling technology to obtain samples randomly. $P_x$ is the negative sample distribution, which is proportional to node $u$'s degree. In this way, we constrain the positive sample intensity to be as large as possible and the negative sample intensity to be as small as possible. 

\subsubsection{Alignment Loss}

Note that we do not utilize the node representations from the temporal module as the final output because they are simple mappings of node origin features. Such node representations are used only to model temporal information as a complement to the structural information. To achieve this goal, we leverage the alignment loss the constrain the temporal intensity $\lambda^T$ and $\lambda^S$ to be as close as possible, thus constraining the temporal information as a complement to the structural information. It means that these two intensities can be compared to each other, thereby guiding the model to strike a balance between different information preferences. Here we utilize the Smooth $L1$ loss to measure it,
\begin{equation}
    L_A = Smooth(\lambda^T, \lambda^S).
    \label{la}
\end{equation}

Denote $\lambda^T - \lambda^S$ as $\Delta\lambda$, the Smooth $L1$ loss can be formulated as follows,
\begin{equation}
    Smooth(\lambda^T, \lambda^S) = 
    \begin{cases}
        \frac{1}{2} (\Delta\lambda)^2,& \text{$|\Delta\lambda| < 1$} \\
        |\Delta\lambda| - \frac{1}{2},& \text{$|\Delta\lambda| \geq 1$}.
    \end{cases}
\end{equation}

Note that both the structural conditional intensity $\lambda^S$ and the temporal conditional intensity $\lambda^T$ are vectors, and when constrained using the Smooth $L$1 loss, it is the values at each position in the intensity vector that are drawn close together.

\subsubsection{Global Loss}

For the global representation and global parameters in this module, we construct a loss function $L_G$ to constrain their variation,
\begin{equation}
    L_G = \log\sigma(- || z_x^t - z_g ||^2 - || z_y^t - z_g ||^2) + ||\alpha||^2 + ||\beta||^2.
    \label{lg}
\end{equation}

Among them, the global representation should be as similar to the node representation as possible to maintain its smoothness, and the global parameter should be as close to $0$ as possible. In this way, we can ensure the global parameter's finely adjusting ability because its values are constrained to transform in a small range.

\subsubsection{Total Loss}
The total loss function contains several parts: the task loss $L_{task}$, the alignment loss $L_A$, and the global loss $L_G$. According to Eq. (\ref{lt1}), (\ref{la}), and (\ref{lg}), the total loss function $L$ can be formally defined as follows,
\begin{equation}
    L = L_{task} + \eta_1 L_A + \eta_2 L_G,
    \label{final loss}
\end{equation}

where $\eta_1$ and $\eta_2$ are learnable parameters used to weigh the constraint of alignment loss and global loss. In general, these two parameters should be set as hyper-parameters and fine-tuned according to the experimental results. But we found that setting it as a learnable parameter was equally effective and more flexible, minimizing the frequency of manual intervention by researchers.

\subsection{Complexity Analysis}

To analysis the time complexity of S2T, we first given its pseudo-code shown in Algorithm. \ref{S2T code}.

\begin{algorithm}[t]
    \caption{S2T procedure}
    \label{S2T code}
    \setstretch{1.1}
    \begin{algorithmic}[1]
        \REQUIRE Temporal graph $G=(V,\ E,\ T,\ X)$.\\
        \ENSURE Node representations.\\
        \STATE {Initialize global representation $z_g$ and parameters;}
        \STATE {Split $G$ in batches;}
        \REPEAT
        \FOR {each $batch$}
        \STATE {Calculate node representations based on Eq. (\ref{GNN});}
        \STATE {Update global representation $z_g$ based on Eq. (\ref{global update});} 
        \STATE {Update node representations based on Eq. (\ref{node emb update});}
        \STATE {Calculate global parameter $\omega_g$ based on Eq. (\ref{omega g});}
        \STATE {Calculate $\lambda^S_{(x,y)}(t)$ based on Eq. (\ref{local intensity});}
        \STATE {Calculate $\lambda^T_{(x,y)}(t)$ based on Eq. (\ref{hawkes lambda});}
        \STATE {Optimize the loss function based on Eq. (\ref{final loss});} 
        \ENDFOR
        \UNTIL{Convergence} 
    \end{algorithmic}
\end{algorithm}

Let $|E|$ be the number of edges, $t$ be the number of epochs, $d$ be the representation size, $l$ be the number of GNN layers, $S$ be the length of the historical neighbor sequence, and $Q$ be the number of negative sample nodes.

According to Algorithm. \ref{S2T code}, we can discuss the time complexity by line:
\begin{enumerate}
    \item Lines 1-2. The complexity of initialization of the parameters depends on the largest size parameter, here it is $O(d^2)$. Splitting graph by batches is equivalent to traversing the graph, whose complexity is denoted as $O(|E|)$.
    
    \item Line 5. The computation of node representation based on GNN is related to the number of layers and the number of neighbors. The complexity of the representation of the previous layer is converted to $O(d^2)$, and the complexity of computing domain information of the current layer is $O(S^2d^2)$. Consider the number of layers, the complexity of this part is $O(lS^2d^2)$.
    
    \item Lines 6-7. For each node, the updating of global representation and node representation have the same complexity, i.e., $O(d)$.
    
    \item Line 8. The complexity of calculating global parameter needs to consider the structure of FiLM layer, which can be denoted as $O(d^2)$.
    
    \item Lines 9-11. The optimization is divided into two steps. Firstly, the results of each part of the loss function are calculated by forward propagation, and then the model parameters are optimized by back propagation. The complexity of forward propagation is $O(QS^3d^3)$, the complexity of back propagation is $O(d^2)$.
    
\end{enumerate}

Considering the number of epochs $t$ and edges $|E|$ outside of the loop, its time complexity can be formalized as follows,
\begin{equation}
    \begin{aligned}
        & O(d^2 + |E| + t|E|(lS^2d^2 + d + d^2 + QS^3d^3 + d^2)) \\& = O(t|E|(lS^2d^2 + QS^3d^3)).
    \end{aligned}
\end{equation}

Because $l, S, Q$ are all small constants, the time complexity of S2T can be simplified as $O(t|E|d^3)$.

\subsection{Discussion}

\subsubsection{Inductive Learning}

S2T can handle new nodes and edges as well, and it is an implicitly inductive model. For a new node-interaction join, we only need to obtain its features and interaction neighbors to generate its node representations from readily available GNNs and global representations. Note that S2T processes the temporal graph data in batches, and each batch of data is equivalent to new nodes and interactions for it, so it is inherently inductive. In fact, almost all temporal models are natural inductive learning models.

\subsubsection{Information Complementary Analysis}

In this part, we discuss the information complementary of the three modules: temporal information modeling, local structural information modeling, and global structural information modeling. Here we measure the modeling scope of different modules by node sequences. Given a node $x$, its one-order neighbors can be defined as $N_x^1$. The temporal module's scope is equal to it, i.e., $S_t = \{N_x^1\}$, because the module only focuses on the one-order neighbor sequence.

The local structural module further pays attention to high-order neighbors based on GNN, thus the number of GNN layers can be used to evaluate the order of neighborhood. In this way, the module's scope can be formulated as $S_l = \{N_x^1 + N_x^2 + \cdots + N_x^l\}$. The global structural module captures information over the whole graph and each node in the graph will be used to update the global representation and parameter, thus the global module's scope can be formulated as $S_g = \{V\}$.

In terms of the modeling scope, the temporal module's scope is contained in the local module, which in turn is contained in the global module, i.e., $\{N_x^1\} \in \{N_x^1 + N_x^2 + \cdots + N_x^l\} \in \{V\}$. And in terms of modeling depth, the temporal graph module digs the deepest information, while the global module digs the shallowest. This combination of models is logical. In a graph, the most likely to influence a node is its first-order neighbors, followed by its higher-order neighbors, and the node is also influenced by the global environment. As the depth of the module modeling decreases, the respective field of S2T is expanding. Furthermore, each module captures information that can be used as a complement to the previous module's information. We will demonstrate the effectiveness of each module individually in the experiments below.

\section{Experiment}
\label{experiment}

\begin{table}[t]
    \centering
    \caption{Description of The Datasets.}
    \label{datasets}
    \begin{threeparttable}
       \resizebox{0.48\textwidth}{10.5mm}{
        \begin{tabular}{c|c c c c c}
            \toprule[2pt]
            Datasets& Wikipedia& CollegeMsg& cit-HepTh& BITotc& Amazon \\
            \midrule[1pt]
            \# Nodes& 8,227& 1,899& 7,557& 5,881& 74,526 \\
            \# Interactions& 157,474& 59,835& 51,315& 35,592& 89,689 \\
            \# Timestamps& 115,858& 50,065&78& 27,487& 5,804 \\
            \# Type& Web& Message& Citation& Bitcoin& Business \\
            \bottomrule[2pt]
        \end{tabular}}
    \end{threeparttable}
\end{table}

\begin{table*}[t]
    \centering
    \caption{Link Prediction Performance. The best results are bolded and the sub-optimal results are underlined.}
    \label{link}
    \begin{threeparttable}
        \resizebox{1\textwidth}{32mm}{
        \begin{tabular}{c | c c | c c | c c | c c | c c}
            \toprule[2pt]
            \multirow{2}{*}{Datasets}& \multicolumn{2}{c |}{Wikipedia}&	\multicolumn{2}{c |}{CollegeMsg}&	\multicolumn{2}{c |}{cit-HepTh}&	\multicolumn{2}{c |}{BITotc}&	\multicolumn{2}{c}{Amazon} \\
            \cmidrule{2-11}
            & ACC& F1& ACC& F1& ACC& F1& ACC& F1& ACC& F1\\
            \midrule[1pt]
            DeepWalk&	65.12$\pm$0.94&	64.25$\pm$1.32& 66.54$\pm$5.36& 67.86$\pm$5.86& 51.55$\pm$0.90&	50.39$\pm$0.98& 52.25$\pm$0.71& 51.99$\pm$1.44& 60.67$\pm$1.86& 64.83$\pm$1.24\\
            
            node2vec&	75.52$\pm$0.58&	75.61$\pm$0.52&	65.82$\pm$4.12& 69.10$\pm$3.50& 65.68$\pm$1.90&	66.13$\pm$2.15& 50.31$\pm$1.12& 57.99$\pm$1.42& 60.00$\pm$3.41& 61.93$\pm$2.53\\
            
            VGAE&	66.35$\pm$1.48&	68.04$\pm$1.18&	65.82$\pm$5.68& 68.73$\pm$4.49& 66.79$\pm$2.58&	67.27$\pm$2.84& 56.81$\pm$1.22& 60.73$\pm$2.40& 57.42$\pm$1.09& 61.83$\pm$1.22\\
            
            GAE&	68.70$\pm$1.34&	69.74$\pm$1.43&	62.54$\pm$5.11& 66.97$\pm$3.22& 69.52$\pm$1.10&	70.28$\pm$1.33& 53.54$\pm$0.78& 56.23$\pm$1.47& 56.34$\pm$1.82& 59.77$\pm$1.54\\
            
            GraphSAGE&	72.32$\pm$1.25&	73.39$\pm$1.25&	58.91$\pm$3.67& 60.45$\pm$4.22& 70.72$\pm$1.96&	71.27$\pm$2.41& 55.39$\pm$0.64& 59.67$\pm$1.62& 63.32$\pm$3.48& 65.54$\pm$2.10\\
            \midrule[0.3pt]
            CTDNE&	60.99$\pm$1.26&	62.71$\pm$1.49&	62.55$\pm$3.67& 65.56$\pm$2.34& 49.42$\pm$1.86&	44.23$\pm$3.92& 60.64$\pm$2.77& 61.28$\pm$1.63& 60.84$\pm$1.55& 62.77$\pm$1.63\\
            
            HTNE&	77.88$\pm$1.56&	78.09$\pm$1.40&	73.82$\pm$5.36& 74.24$\pm$5.36& 66.70$\pm$1.80&	67.47$\pm$1.16& 69.12$\pm$0.88& 71.45$\pm$1.83& \underline{80.62$\pm$2.47}& \underline{82.03$\pm$3.38}\\
            
            MMDNE&	79.76$\pm$0.89&	79.87$\pm$0.95&	73.82$\pm$5.36& 74.10$\pm$3.70& 66.28$\pm$3.87&	66.70$\pm$3.39& 65.56$\pm$1.55& 69.33$\pm$0.87& 64.94$\pm$2.01& 67.73$\pm$1.42\\
            
            EvolveGCN&	71.20$\pm$0.88&	73.43$\pm$0.51&	63.27$\pm$4.42& 65.44$\pm$4.72& 61.57$\pm$1.42&	62.42$\pm$1.54& 69.79$\pm$0.64& 72.79$\pm$1.31& 69.59$\pm$1.32& 71.43$\pm$2.87\\
            
            TGN&	73.89$\pm$1.42&	80.64$\pm$2.79&	66.13$\pm$3.58& 69.84$\pm$4.49& 69.54$\pm$0.98&	82.44$\pm$0.73& \underline{75.12$\pm$2.35}& \underline{75.97$\pm$1.48}& 77.72$\pm$1.96& 78.83$\pm$1.87\\
            
            TGAT&	76.45$\pm$0.91&	76.99$\pm$1.16&	58.18$\pm$4.78& 57.23$\pm$7.57& 78.02$\pm$1.93&	78.52$\pm$1.61& 73.62$\pm$0.86& 71.94$\pm$2.55& 70.42$\pm$3.78& 73.59$\pm$1.66\\
            
            MNCI&	78.86$\pm$1.93&	74.35$\pm$1.47&	66.34$\pm$2.18& 62.66$\pm$3.22& 73.53$\pm$2.57&	72.84$\pm$4.31& 70.53$\pm$1.32& 69.89$\pm$1.78& 73.03$\pm$2.52& 72.34$\pm$2.79\\
            
            TREND& \underline{83.75$\pm$1.19}& \underline{83.86$\pm$1.24}&	\underline{74.55$\pm$1.95}& \underline{75.64$\pm$2.09}& \underline{80.37$\pm$2.08}& \underline{81.13$\pm$1.92}& 73.73$\pm$2.48& 75.14$\pm$1.62& 75.69$\pm$2.87& 76.06$\pm$1.56\\
            
            \midrule[0.3pt]
            
            S2T&	\textbf{88.01$\pm$1.04}&	\textbf{87.92$\pm$0.97}&	\textbf{76.81$\pm$2.03}& \textbf{77.25$\pm$2.16}& \textbf{88.83$\pm$1.64}& \textbf{89.04$\pm$1.33}& \textbf{78.81$\pm$1.27}& \textbf{79.74$\pm$1.56}& \textbf{88.42$\pm$2.21}& \textbf{88.54$\pm$1.73}\\

            (improv.)& (+5.08\%)& (+4.84\%)&	(+3.03\%)& (+2.12\%)& (+10.52\%)& (+8.00\%)& (+4.91\%)& (+4.96\%)& (+9.67\%)& (+7.93\%)\\
            
            \bottomrule[2pt]
        \end{tabular}}
    \end{threeparttable}
\end{table*}

\subsection{Datasets}

The description of datasets is presented in Table \ref{datasets}. \textbf{Wikipedia} \cite{kumar2019predicting} is a web graph, which contains the behavior of people editing web pages on Wikipedia, and each edit operation is regarded as an interaction. \textbf{CollegeMsg} \cite{panzarasa2009patterns} is an online social graph where one message between two users is considered as an interaction. \textbf{cit-HepTh} \cite{leskovec2005graphs} is a academic graph that includes the citation records of papers in the high energy physics theory field. \textbf{BITotc} \cite{kumar2018rev2} is a dataset of bitcoin transactions in which users make repeated transactions on the platform. \textbf{Amazon} \cite{ni2019justifying} is an interactive record dataset of user reviews of magazines on Amazon website.

\subsection{Baselines}

In this part, multiple methods are introduced to compare with S2T. We divide them into two parts: static graph methods and temporal graph methods.

(1) Static graph-based methods: \textbf{DeepWalk} \cite{perozzi2014deepwalk} is a classic work in this field, which performs random walks over the graph to learn node embeddings. \textbf{node2vec} \cite{grover2016node2vec} conducts random walks on the graph using breadth-first and depth-first strategies to balance neighborhood information of different orders. \textbf{VGAE} and \textbf{GAE} \cite{kipf2016variational} migrates variational auto-encoders to graph data and use encoder-decoder module to reconstruct graph information. \textbf{GraphSAGE} \cite{hamilton2017inductive} learns an aggregation function to sample and combine features from a node's local neighborhood.

(2) Temporal graph-based methods: \textbf{CTDNE} \cite{nguyen2018continuous} performs random walk on graphs to model temporal ordered sequences of node walks. \textbf{HTNE} \cite{zuo2018embedding} is the first to utilize the Hawkes process to model node influence on temporal graphs. \textbf{MMDNE} \cite{lu2019temporal} models graph evolution over time from both macro and micro perspectives. \textbf{EvolveGCN} \cite{pareja2020evolvegcn} uses the RNN model to update the parameters of GCN for future snapshots. \textbf{TGAT} \cite{xu2020inductive} replaces traditional modeling form of self-attention with interaction temporal encoding. \textbf{MNCI} \cite{liu2021inductive} mines community and neighborhood influences to generate node representations. \textbf{TREND} \cite{wen2022trend} replaces the Hawkes process with GNN to model temporal information.

\subsection{Experiment Settings}

In the hyper-parameter settings, we select Adam \cite{kingma2014adam} optimizer with a learning rate $0.001$. The embedding dimension size $d$, the batch size $b$, the negative sampling number $Q$, and the historical sequence length $S$ are set to 128, 128, 1, and 10, respectively. We present the parameter sensitivity analysis on the effect of the hyper-parameters $Q$ and $S$ in Sect. \ref{para}. For the baseline methods, we keep their default parameter settings. 

To evaluate these methods' performance, we conduct link prediction as the basic task. In addition, we further discuss the effect of several parameters and modules on performance through ablation study, parameter sensitivity analysis, loss convergence analysis, complexity comparison, and robustness analysis.

\begin{figure}[t]
    \centering
    \includegraphics[width=0.48\textwidth]{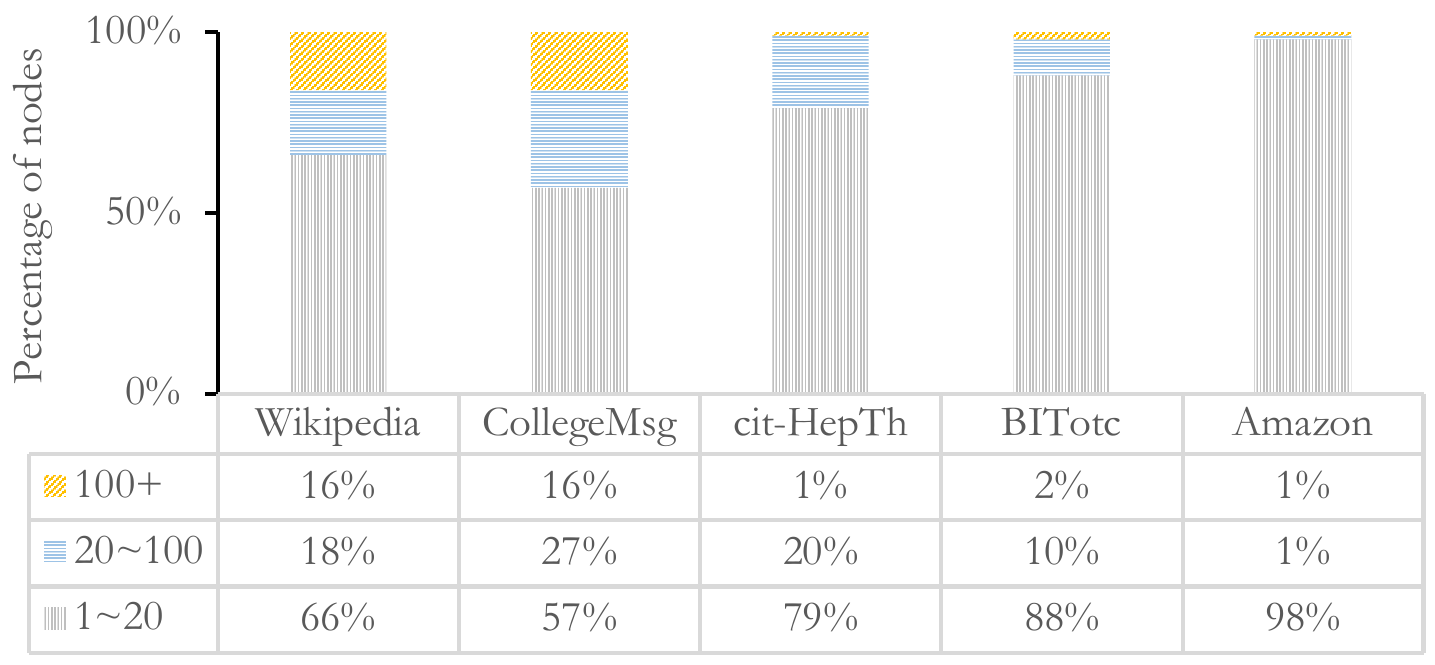}
    \caption{Distribution of Node Degree.}
    \label{node dis}
\end{figure}

\subsection{Link Prediction Results}
\label{linkpre}

In this part, we compare S2T with multiple competitors on the link prediction task and divide the dataset into a training set and a test set in chronological order of 80\% and 20\%. After split the dataset, we first train model on the training set and then conduct link predictions on the test set. In the test set, for each interaction, we define it as positive pair and random sample a negative pair (i.e., two nodes have never interacted with each other). After an equal number of positive and negative sample pairs are generated, we use a logistic regression function to determine the positives and negatives of each pair and compare them with the true results. We leverage the Accuracy (ACC) and F1-Score (F1) as performance metrics. In Table \ref{link}, the proposed S2T achieves the best performances compared with various existing baselines on all datasets.

In these datasets, S2T obtains the best improvement on cit-HepTh. We argue that this phenomenon is related to the distribution of node degrees on different datasets. Thus we provide pie charts of degree distributions to explain this problem. In Figure \ref{node dis}, the number of nodes with different degrees is given. We simply define nodes with a degree between $1$ and $20$ as low-active nodes (i.e., long-tail nodes) and nodes with a degree above $100$ as high-active nodes, then can find that the number of low-active nodes accounts for a higher percentage than the sum of other two categories. This is in line with the phenomenon we pointed out above that long-tail nodes are the most common category of nodes in the graph.

Here we give the number of long-tail nodes on datasets: Wikipedia (5045, 65.69\%), CollegeMsg (1082, 56.97\%), cit-HepTh (5999, 79.17\%), BITotc (5211, 88.61\%), and Amazon (73897, 99.15\%). Moreover, in conjunction with Table \ref{link}, if a dataset has a larger proportion of long-tailed nodes, the more our global module works, and thus the better S2T improves on that dataset. The experimental results and data analysis nicely corroborates the effectiveness of our proposed global module for information enhancement of long-tail nodes.

As mentioned above, the baselines include two parts: static methods and temporal methods. According to the link prediction result, most of the temporal methods achieve better performance than static methods, which means that the temporal information in node interactions is important. Compared to HTNE, which models temporal information with the Hawkes process, and TREND, which models structural information with GNN, our method S2T achieves better performance by combining both temporal and structural information. This indicates that the alignment loss can constrain S2T to capture the two different types of information effectively.

\begin{figure*}[htbp]
    \centering
    \begin{minipage}[t]{0.3\textwidth}
        \includegraphics[width=1\textwidth]{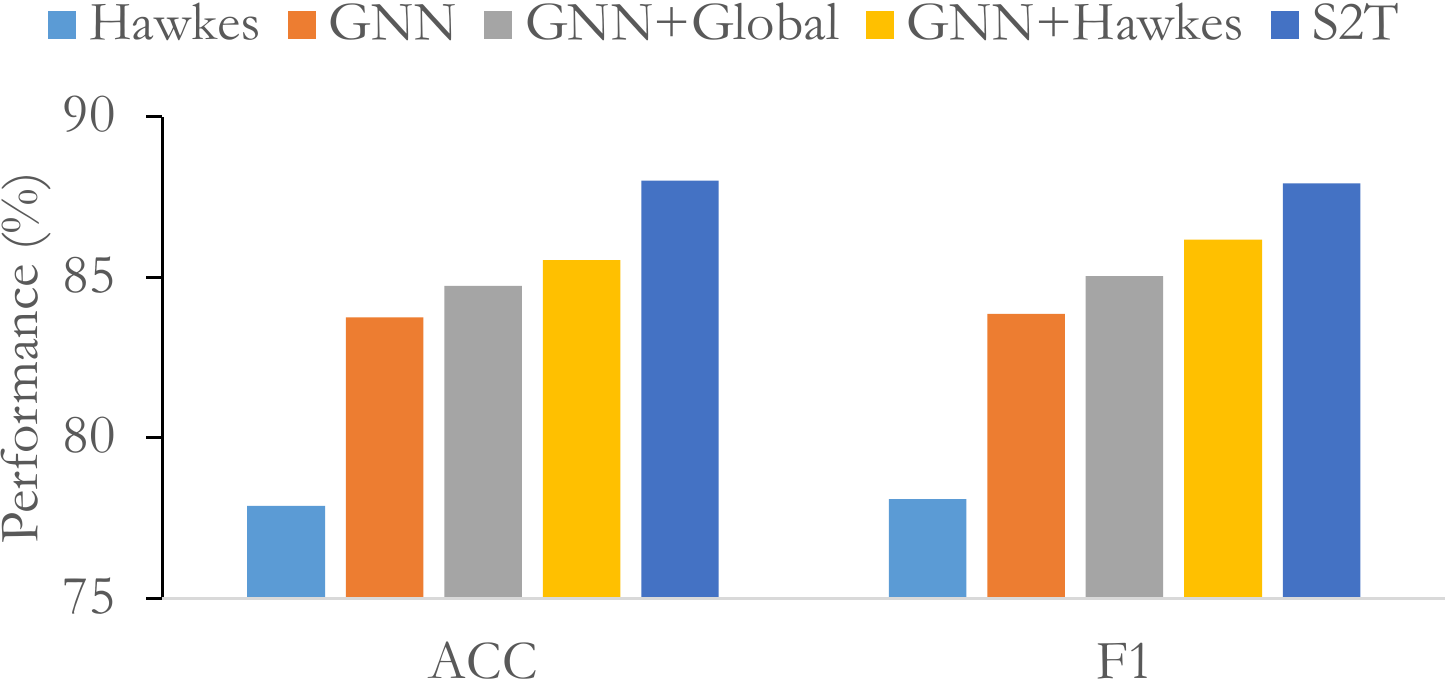}
        \centerline{(a) Wikipedia}
        \newline
    \end{minipage}%
    \hspace{3mm}
    \begin{minipage}[t]{0.3\textwidth}
        \includegraphics[width=1\textwidth]{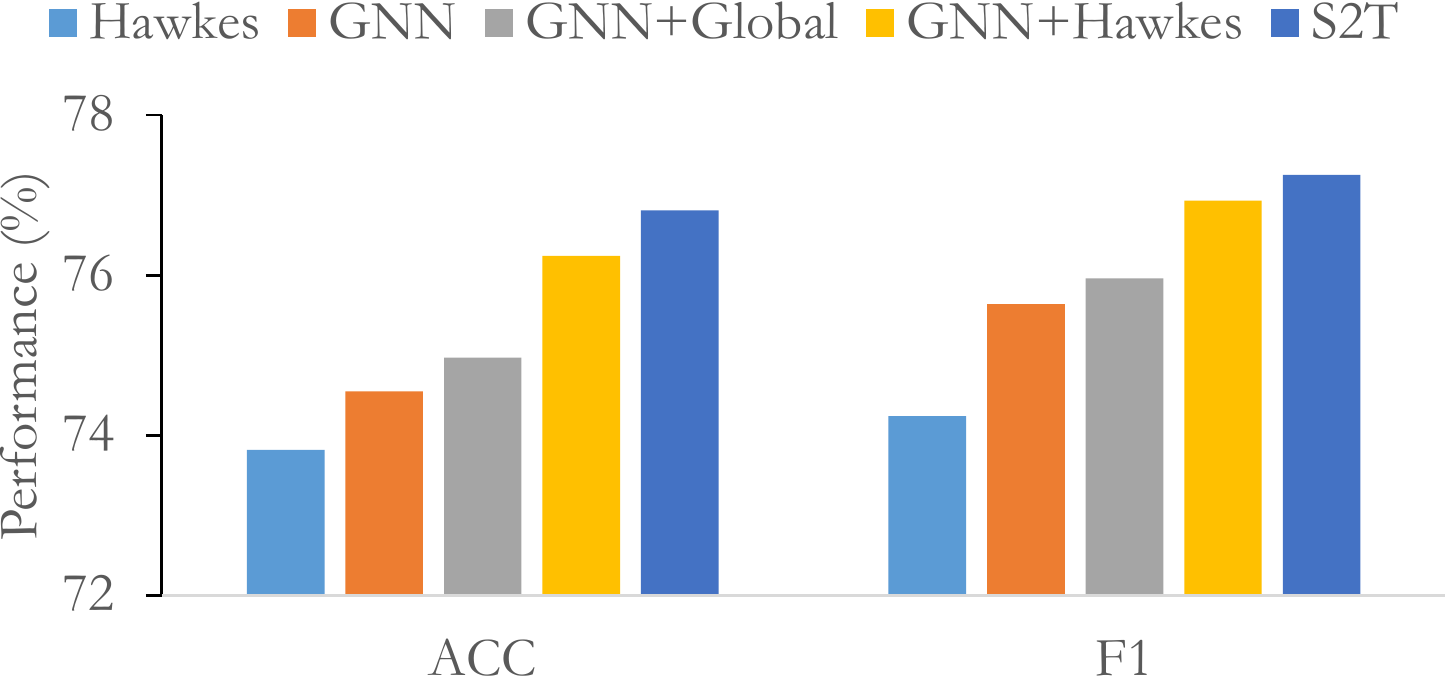}
        \centerline{(b) CollegeMsg}
        \newline
    \end{minipage}%
    \hspace{3mm}
    \begin{minipage}[t]{0.3\textwidth}
        \includegraphics[width=1\textwidth]{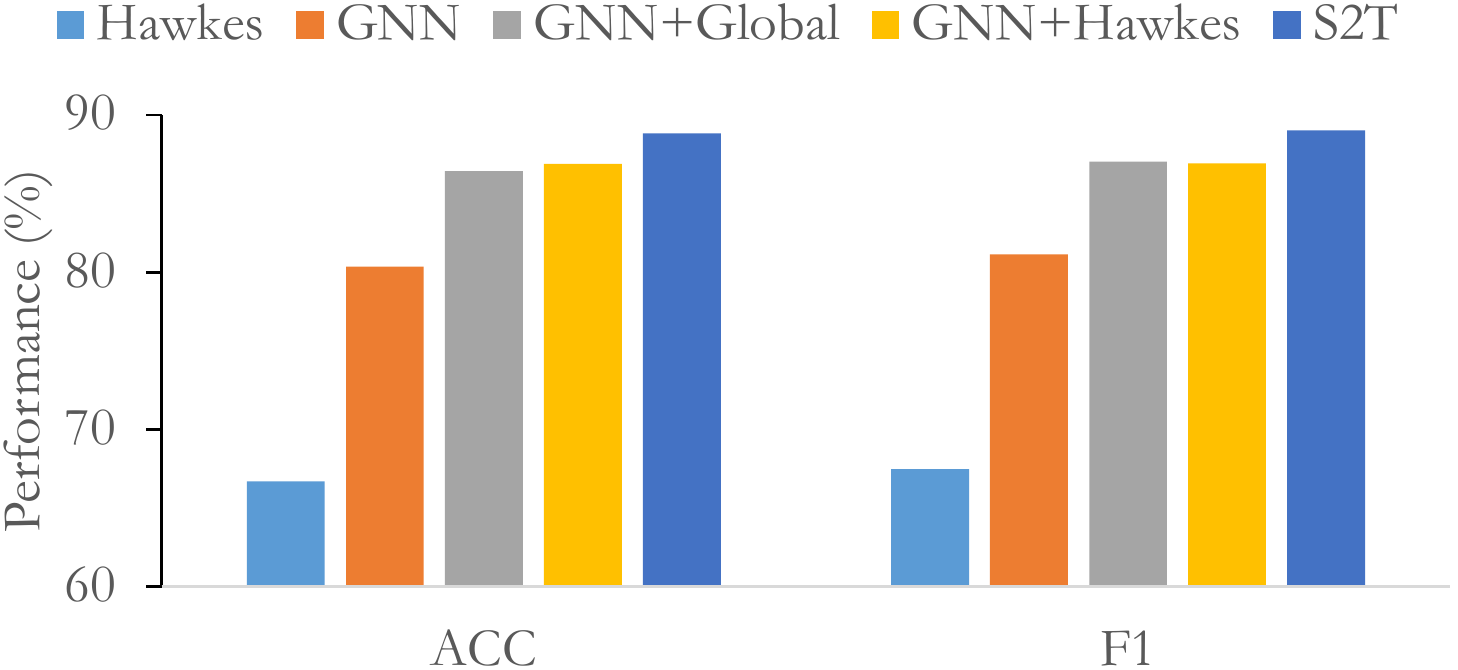}
        \centerline{(c) cit-HepTh}
        \newline
    \end{minipage}%
    \newline
    \begin{minipage}[t]{0.3\textwidth}
        \includegraphics[width=1\textwidth]{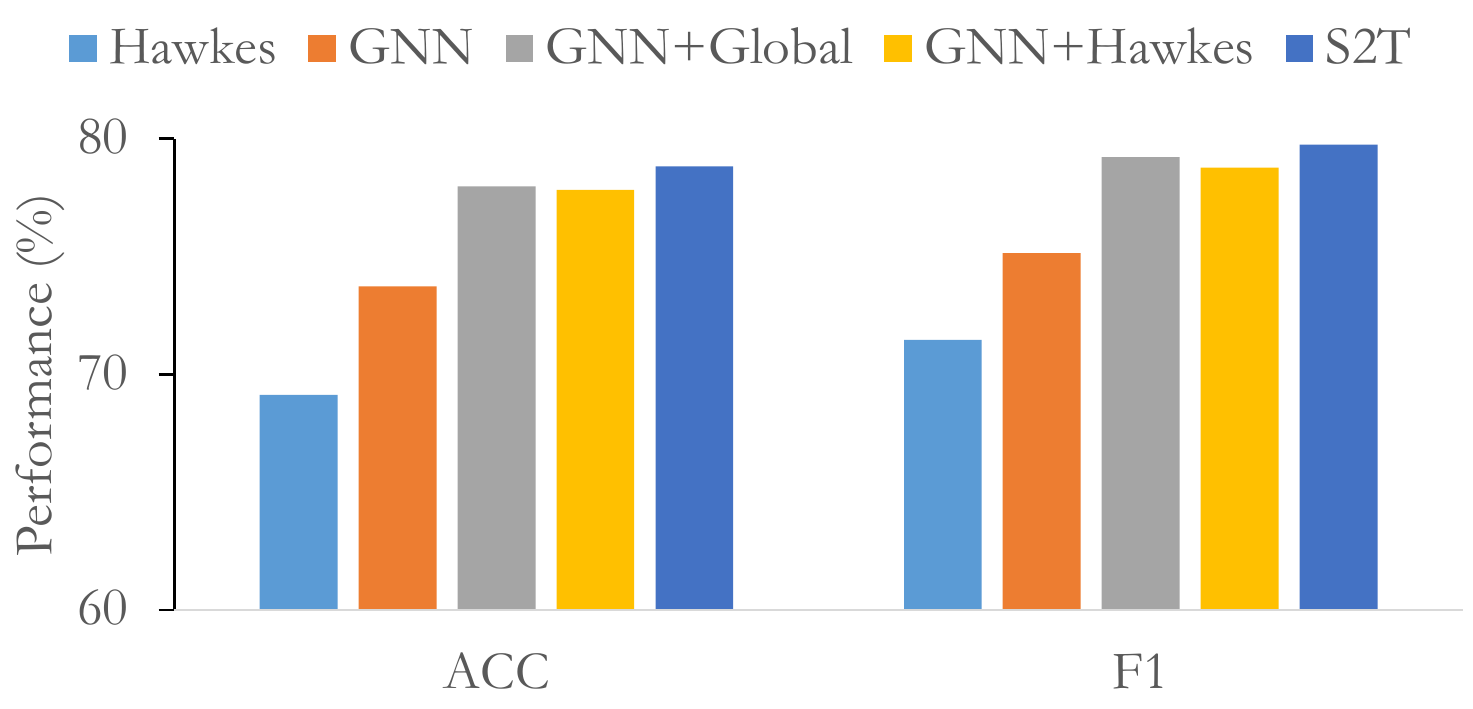}
        \centerline{(d) BITotc}
        \newline
    \end{minipage}%
    \hspace{3mm}
    \begin{minipage}[t]{0.3\textwidth}
        \includegraphics[width=1\textwidth]{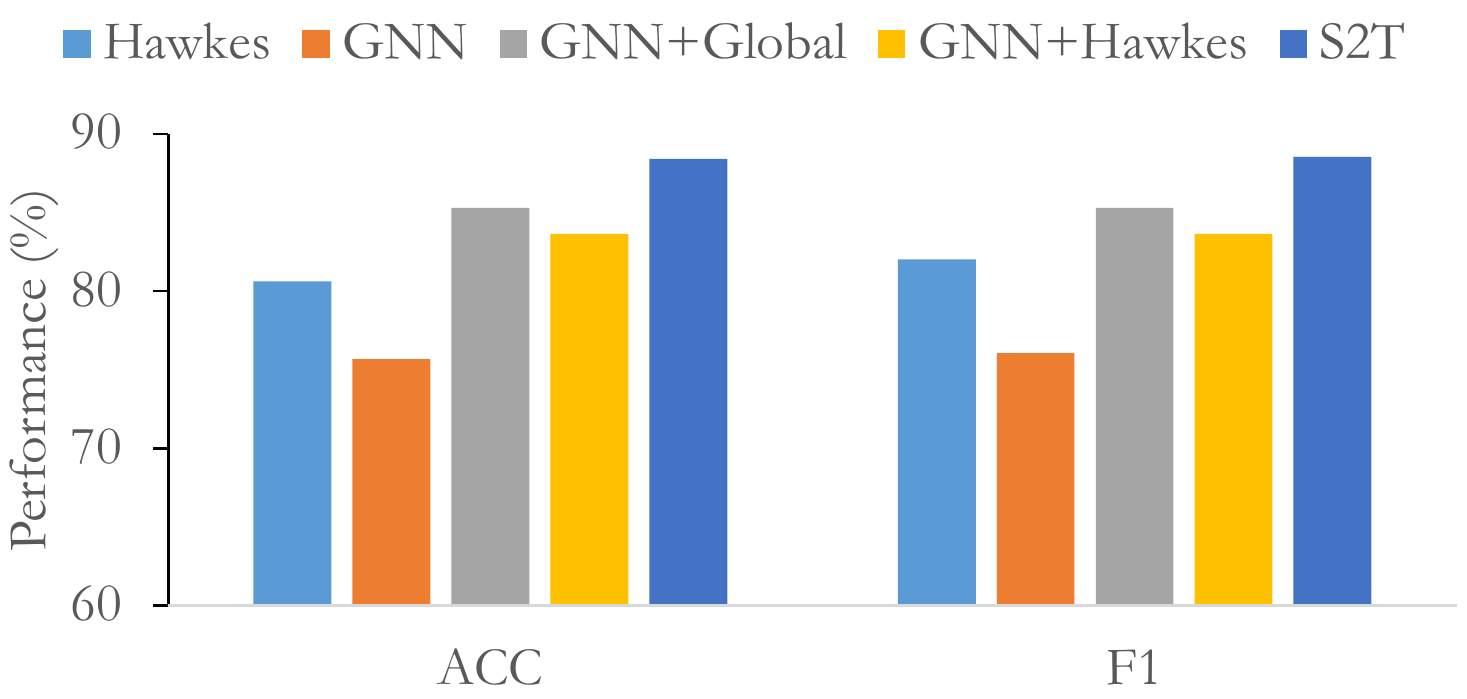}
        \centerline{(e) Amazon}
        \newline
    \end{minipage}%
    \caption{Ablation Study on all Datasets.}
    \label{ablation}
\end{figure*}

\subsection{Ablation Study}

Note that in the proposed S2T, we introduce the temporal information modeling module based on the Hawkes process and the structural information modeling module. The structural module contains a local module based on GNN and a global module. In this part, we will discuss different modules' effects on performance.

More specially, we select five module combinations: (1) only temporal information modeling (i.e., \textbf{Hawkes} module); (2) only local structural information modeling (i.e., \textbf{GNN} module); (3) both local and global structural information modeling (i.e., \textbf{GNN+Global}); (4) align temporal information with local structural neighborhood (i.e., \textbf{GNN+Hawkes}); (5) the final model (i.e., \textbf{S2T}).

As shown in Figure \ref{ablation}, we can find that both Hawkes and GNN modules can only achieve sub-optimal performance. If the two modules are combined, the result of GNN+Hawkes module is significantly improved, which demonstrates the effect of the proposed alignment loss.

Furthermore, when the GNN module incorporates global information, the performance of GNN+Global module is also further improved. By comparing module GNN and GNN+Global, the average magnitude of improvement on all different datasets is 0.49\% on CollegeMsg, 1.28\% on Wikipedia, and 7.40\% on cit-HepTh, which is consistent with the ranking of the long-tailed node proportion in Figure \ref{node dis}. It means that the datasets with more long-tail nodes have a larger performance improvement, which proves that our proposed global module is effective in enhancing long-tail nodes.

\begin{figure*}[t]
    \centering
    \begin{minipage}[t]{0.3\textwidth}
        \includegraphics[width=1\textwidth]{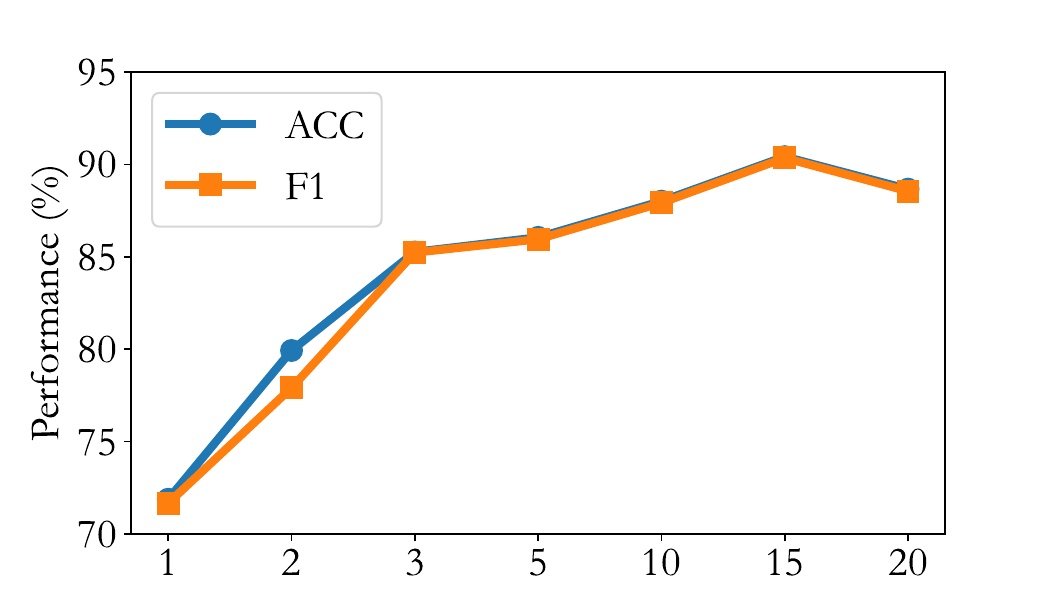}
        \centerline{(a) Wikipedia}
    \end{minipage}%
    \begin{minipage}[t]{0.3\textwidth}
        \includegraphics[width=1\textwidth]{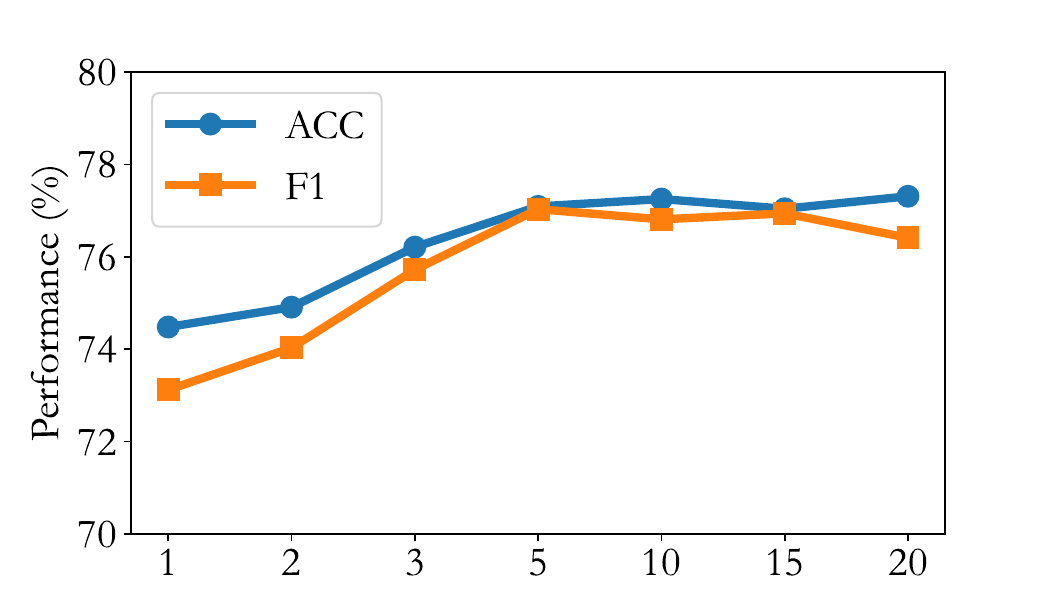}
        \centerline{(b) CollegeMsg}
    \end{minipage}%
    \begin{minipage}[t]{0.3\textwidth}
        \includegraphics[width=1\textwidth]{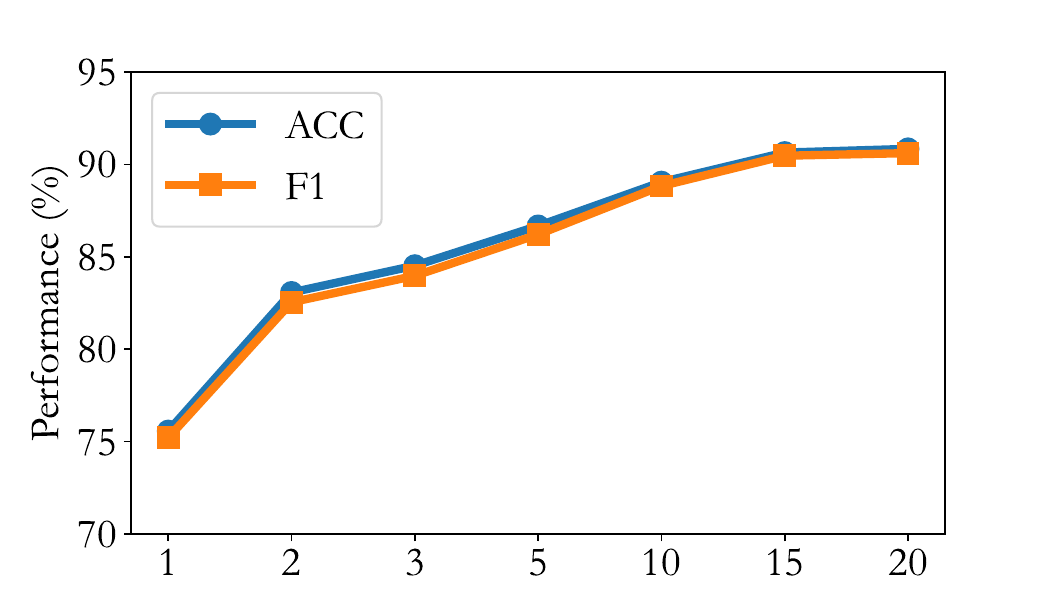}
        \centerline{(c) cit-HepTh}
    \end{minipage}%
    \caption{Parameter Sensitivity of Historical Sequence Length.}
    \label{his}
\end{figure*}

\begin{figure*}[t]
    \centering
    \begin{minipage}[t]{0.3\textwidth}
        \includegraphics[width=1\textwidth]{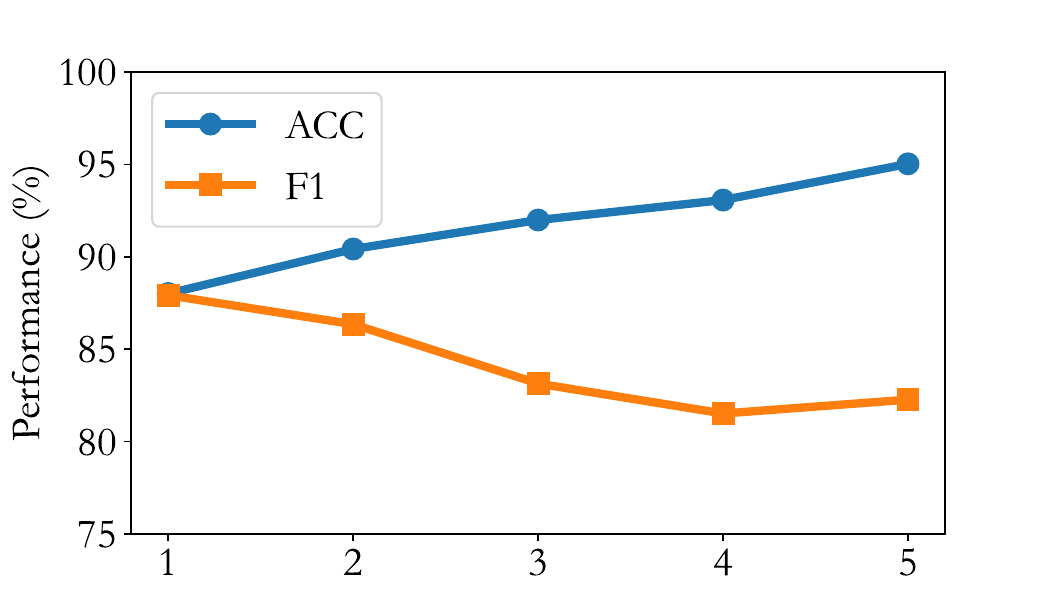}
        \centerline{(a) Wikipedia}
    \end{minipage}%
    \begin{minipage}[t]{0.3\textwidth}
        \includegraphics[width=1\textwidth]{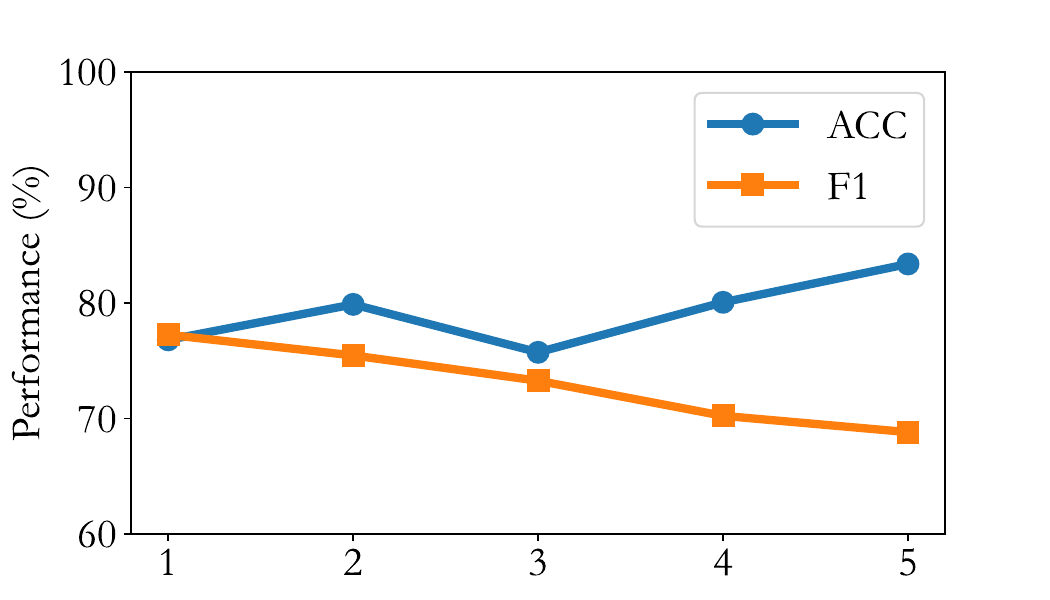}
        \centerline{(b) CollegeMsg}
    \end{minipage}%
    \begin{minipage}[t]{0.3\textwidth}
        \includegraphics[width=1\textwidth]{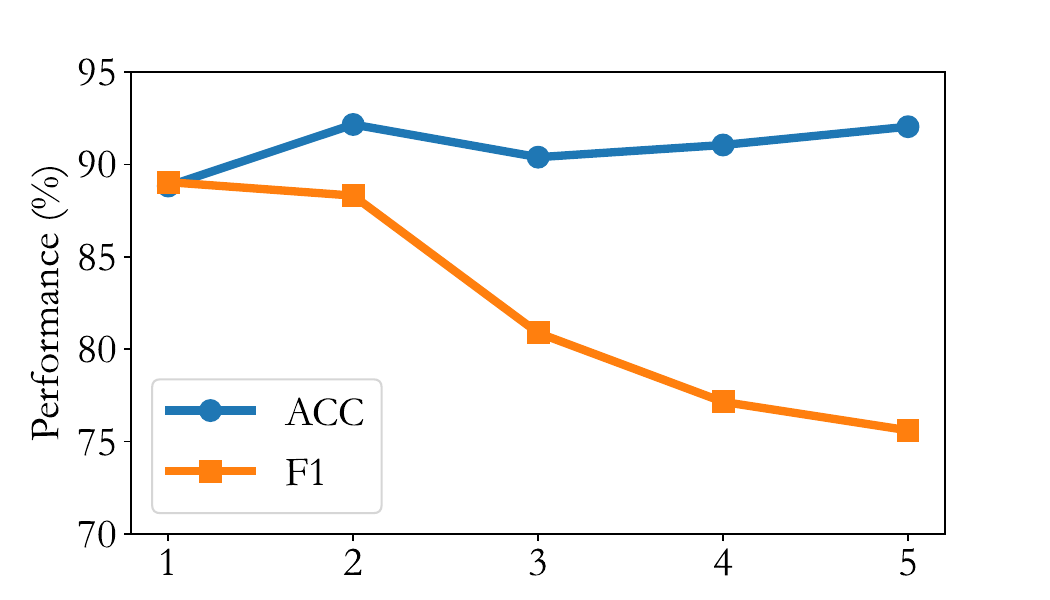}
        \centerline{(c) cit-HepTh}
    \end{minipage}%
    \caption{Parameter Sensitivity of Negative Sample Number.}
    \label{neg}
\end{figure*}

\subsection{Parameter Sensitivity Analysis}
\label{para}

\subsubsection{Length of Historical Neighbor Sequence}

In a temporal graph, node neighbors are fed into the model in batches in the form of interaction sequences. But in actual training, if we obtain all of its neighbors for each node, the computational pattern of each batch can not be fixed, which brings great computational inconvenience. To maintain the convenient calculation of batch training, it is hard for the model to obtain multiple neighbor sequences with different lengths.

According to Figure \ref{node dis}, most nodes have few neighbors, especially in the first half of the time zone. Referencing previous works \cite{zuo2018embedding, hu2020graph, liu2022embtemporal, wen2022trend, xu2020inductive}, many temporal graph methods choose to fix the sequence length $S$ of neighbor sequence and obtain each node's latest $S$ neighbors instead of saving full nodes. If a node doesn't have enough neighbors at a certain timestamp, we mask the empty positions. Thus we need to discuss a question, how do different values of $S$ influence performance?

As shown in Figure \ref{his}, with the change of $S$, the model performance can achieve better results when $S$ is taken as $10/15/20$. In particular, the optimal value of $S$ is taken differently on different datasets. On Wikipedia and cit-HepTh, the optimal value of $S$ are $15$ and $20$, respectively. But on the CollegeMsg dataset, when we select $S$ as $20$, the ACC and F1 performance show a large deviation. In contrast, the two results appear more balanced when $S$ is taken to be $10$ or $15$. For this phenomenon, we argue that with the continuous increase of $S$ ($0-15$), the model can capture more and more neighbor information. But after that, when $S$ continues to increase, too many unnecessary neighbor nodes will be added. These neighbors usually interact earlier, thus the information contained in their interaction can hardly be used as an effective reference for future prediction.

In addition, too many neighbors will increase the amount of computation. Therefore, in the real training, although the performance is better when $S$ is $15$, we default $S$ to $10$ as the hyper-parameter value for the convenience of calculation.

\subsubsection{Negative Sample Number}

The negative sample number $Q$ is a hyper-parameter utilized to control how many negative pairs are generated to the link prediction task loss in Eq. (\ref{lt1}). As shown in Figure \ref{neg}, we can find that with the increase of the negative sample numbers, although the ACC performance increases, the F1 score decreases. It means that an increase in the number of negative samples will lead to an imbalance in the proportion of positive and negative samples in the test, resulting in the above phenomenon. Thus on all datasets, it is robust to select one negative sample pair to benchmark one positive pair.

\subsection{Convergence of Loss}

\begin{figure}[t]
    \centering
    \includegraphics[width=0.4\textwidth]{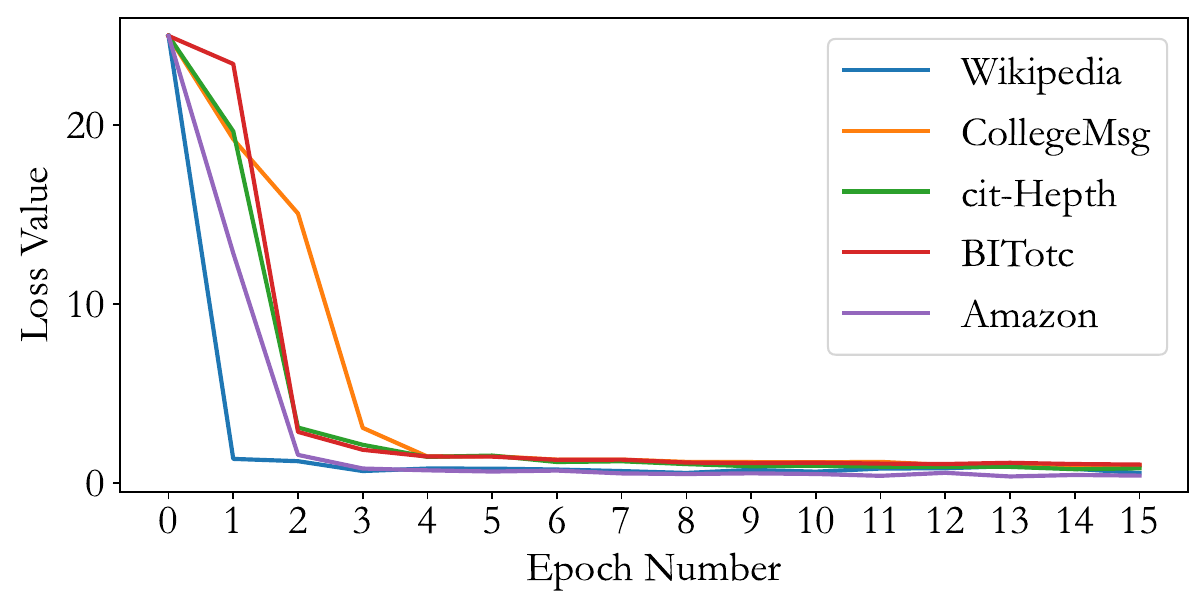}
    \caption{Convergence of Loss.}
    \label{Convergence of Loss}
\end{figure}

As shown in Figure \ref{Convergence of Loss}, in all datasets, the loss values of S2T can achieve convergence after a few epochs. By comparing the convergence laws of the datasets, we find that the dataset with more nodes has a faster convergence speed of the corresponding loss value. It means that more node samples are provided in each training so that the model can learn better.

Combined with the above discussion, S2T has time complexity of $O(t|E|d^3)$ and can converge quickly with a small amount of epoch training, which means that S2T can be more adaptable to large-scale data.

\subsection{Complexity Comparison}

\begin{figure}[t]
    \centering
    \begin{minipage}[t]{0.24\textwidth}
        \includegraphics[width=1\textwidth]{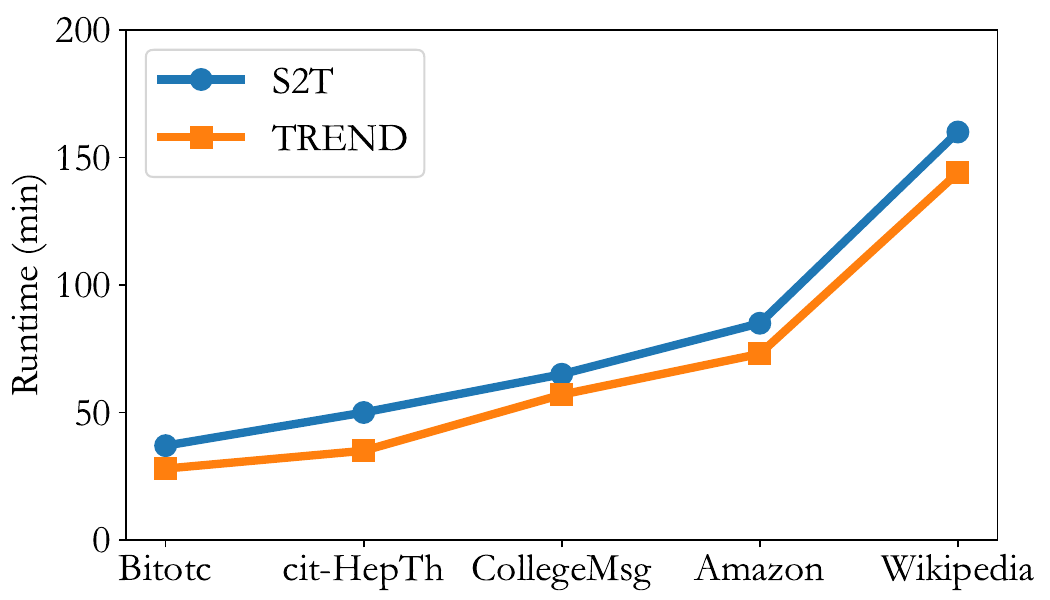}
        \centerline{(a) Runtime}
    \end{minipage}%
    \begin{minipage}[t]{0.24\textwidth}
        \includegraphics[width=1\textwidth]{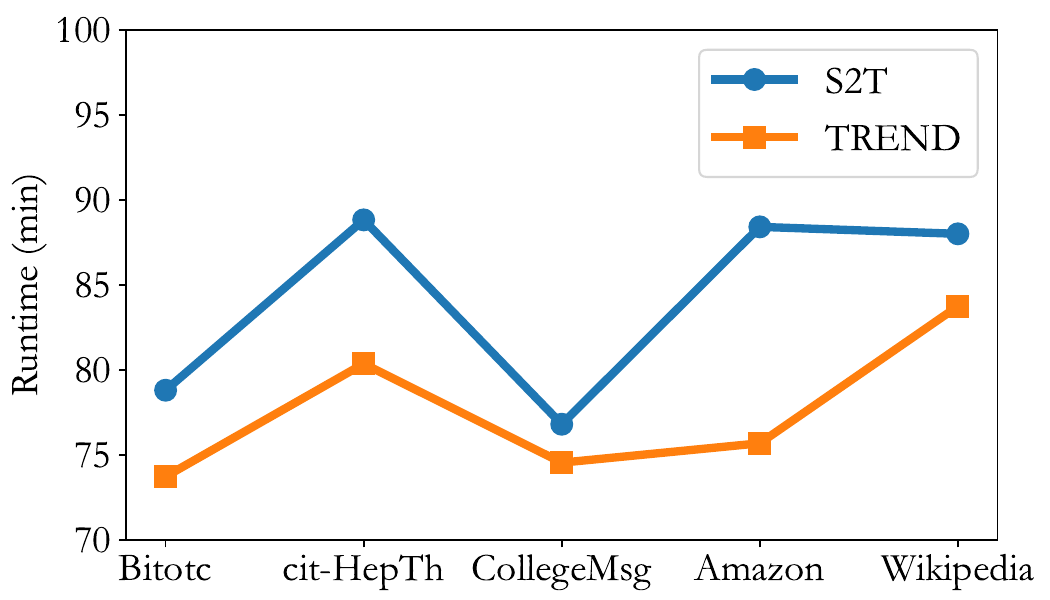}
        \centerline{(b) Performance}
    \end{minipage}%
    \caption{Comparison Between S2T and TREND.}
\label{Comparison}
\end{figure}

Here, we further discuss the additional time complexity that comes with considering temporal information using the Hawkes process. As can be seen from the Algorithm \ref{S2T code}, the complexity of the time information is mainly focused on (1) calculating the temporal information intensity and (2) intensity alignment.

For the first part, as shown in Line 10, we can calculate the complexity as $O(S^3d^2)$. And the complexity of second part is $O(Qd)$. Without considering the time intensity, the complexity of the original method can be simplified to $O(d^2 + |E| + t|E|(lS^2d^2 + d + d^2 +S^3d^3 + d^2))$, i.e., $O(t|E|(lS^2d^2 + S^3d^3))$. By comparison we can see that just a smaller constant $Q$ has been omitted, which does not constitute a major complexity. That is, even after removing the time intensity, the core complexity of the method is still $O(t|E|d^3)$, which remains unchanged.

Certainly, we should acknowledge that although the theoretical time complexity does not change, there is a difference in the actual running time. This is because the model needs to be optimized for more steps in the back-propagation process, but we think this increase in time is acceptable. As shown in Figure \ref{Comparison}, we compared the running time and experimental performance of the two models S2T and TREND. In fact, the time increase of S2T compared to TREND is limited, and we believe that it is meaningful to exchange a small time sacrifice for a performance improvement.

\subsection{Robustness Analysis}

\begin{figure}[t]
    \centering
    \begin{minipage}[t]{0.24\textwidth}
        \includegraphics[width=1\textwidth]{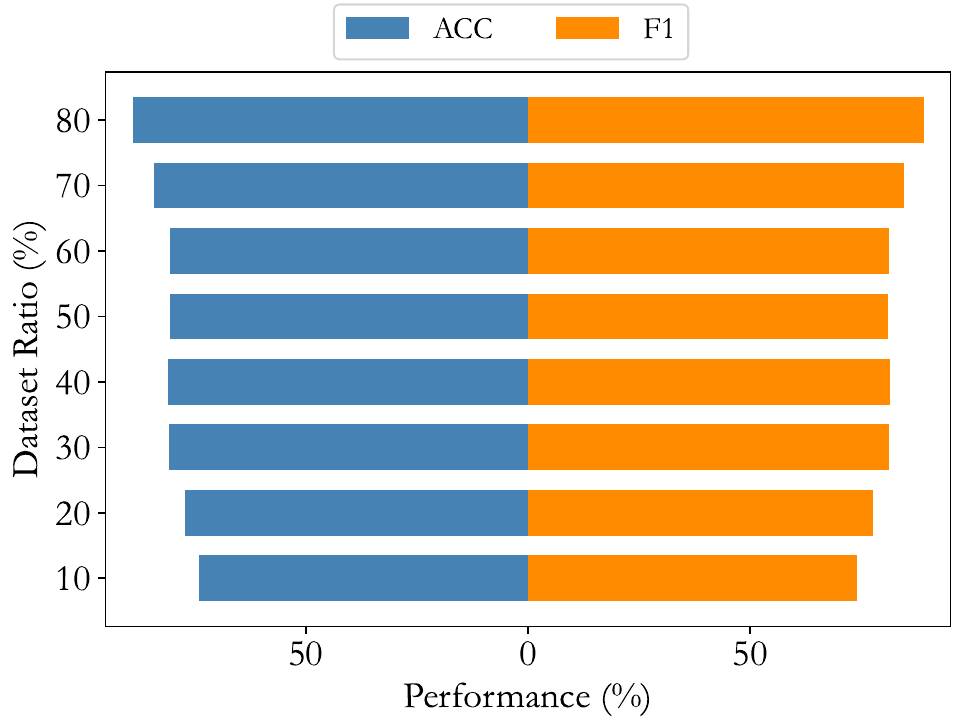}
        \centerline{(a) cit-HepTh}
    \end{minipage}%
    \begin{minipage}[t]{0.24\textwidth}
        \includegraphics[width=1\textwidth]{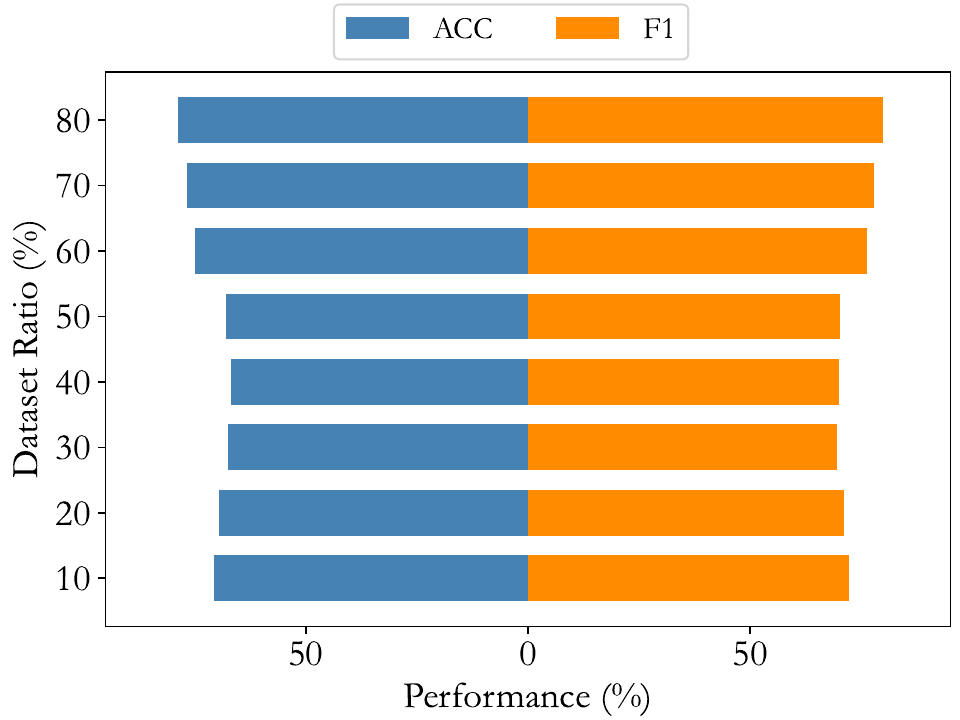}
        \centerline{(b) BITotc}
    \end{minipage}%
    \caption{Performance of Different Dataset Ratios.}
    \label{incomplete}
\end{figure}

\begin{figure}[t]
    \centering
    \includegraphics[width=0.45\textwidth]{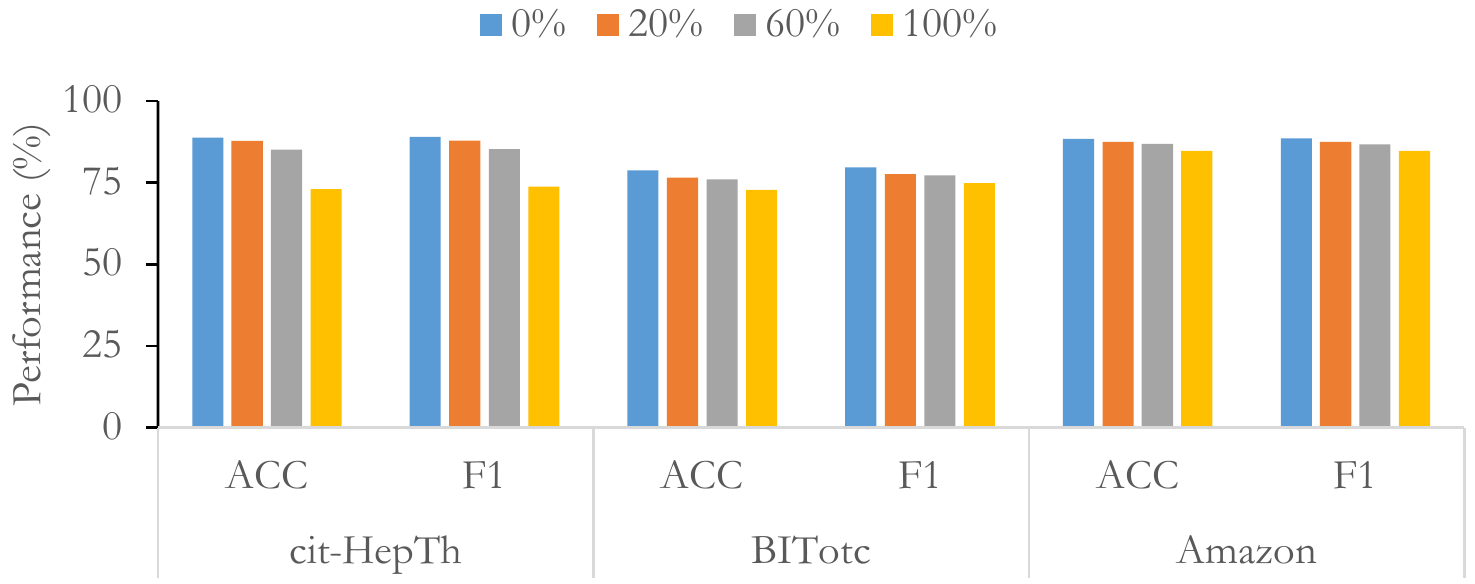}
    \caption{Performance of Different Noisy Ratios.}
    \label{noisy}
\end{figure}

Here, we conduct experiments on incomplete graphs and noisy graphs, respectively, to examine the resilience of S2T.

Regarding the analysis of incomplete graphs, we sampled the training set at various proportions, commencing at 10\% and increasing by 10\% increments until reaching the original proportion of the training set (i.e., 80\%). Figure \ref{incomplete} showcases the performed experiments on training sets of different proportions, alongside the reported performances. Observing the experimental results, it becomes apparent that while scaling down the dataset leads to a decline in performance, the overall decrease remains limited. This implies that our model retains the ability to effectively learn latent data distributions even with smaller data sizes, thereby demonstrating its robustness in handling incomplete graphs.

Concerning the analysis of noisy graphs, we employed different ratios to introduce noise and devised two strategies for its incorporation. The first strategy involved adding interactions that are absent in the dataset (where nodes exist but interactions do not), while the second strategy entailed modifying the temporal information of existing interactions. If we claim to have added 20\% noise, then each of the two strategies contributed to a 10\% increase in noisy data.

As depicted in Figure \ref{noisy}, it can be observed that as the noisy ratio reaches 100\%, there is a noticeable decline in performance; however, it remains relatively competitive. Conversely, smaller noise ratios have a limited impact on performance. This suggests that our proposed method, S2T, excels in extracting latent laws governing node interactions rather than optimizing node features. We firmly believe that the extraction of node interaction laws is a crucial capability that enhances the generality and robustness of S2T.

In summary, through various experimental setups, we can conclude that S2T maintains a favorable level of performance, characterized by heightened robustness, when confronted with complex data distributions.

\section{Conclusion}
\label{conclusion}

We propose a self-supervised graph learning method S2T, by extracting both temporal and structural information to learn more informative node representations. The alignment loss is introduced to narrow the gap between temporal and structural intensities, which can encourage the model to learn both valid temporal and structural information. We also construct a global module to enhance the long-tail nodes' information. Experiments on several datasets prove the proposed S2T achieves the best performance in all baseline methods. In the future, we will try to construct a more general framework to combine multi-modal information.

\ifCLASSOPTIONcompsoc
\section*{Acknowledgments}
\else
\section*{Acknowledgment}
\fi

This work was supported by the National Key R\&D Program of China (no. 2020AAA0107100), the National Natural Science Foundation of China (no. 62325604, 62276271), and a public service platform for artificial intelligence screening and auxiliary diagnosis for the medical and health industry, Ministry of Industry and Information Technology of the People's Republic of China (2020-0103-3-1).

\balance

\bibliographystyle{IEEEtran}
\bibliography{tnnls}

\begin{thebibliography}{10}
\providecommand{\url}[1]{#1}
\csname url@samestyle\endcsname
\providecommand{\newblock}{\relax}
\providecommand{\bibinfo}[2]{#2}
\providecommand{\BIBentrySTDinterwordspacing}{\spaceskip=0pt\relax}
\providecommand{\BIBentryALTinterwordstretchfactor}{4}
\providecommand{\BIBentryALTinterwordspacing}{\spaceskip=\fontdimen2\font plus
\BIBentryALTinterwordstretchfactor\fontdimen3\font minus
  \fontdimen4\font\relax}
\providecommand{\BIBforeignlanguage}[2]{{%
\expandafter\ifx\csname l@#1\endcsname\relax
\typeout{** WARNING: IEEEtran.bst: No hyphenation pattern has been}%
\typeout{** loaded for the language `#1'. Using the pattern for}%
\typeout{** the default language instead.}%
\else
\language=\csname l@#1\endcsname
\fi
#2}}
\providecommand{\BIBdecl}{\relax}
\BIBdecl

\bibitem{cui2018survey}
P.~Cui, X.~Wang, J.~Pei, and W.~Zhu, ``A survey on network embedding,''
  \emph{TKDE}, 2018.

\bibitem{wu2020comprehensive}
Z.~Wu, S.~Pan, F.~Chen, G.~Long, C.~Zhang, and S.~Y. Philip, ``A comprehensive
  survey on graph neural networks,'' \emph{IEEE transactions on neural networks
  and learning systems}, vol.~32, no.~1, pp. 4--24, 2020.

\bibitem{qiu2018network}
J.~Qiu, Y.~Dong, H.~Ma, J.~Li, K.~Wang, and J.~Tang, ``Network embedding as
  matrix factorization: Unifying deepwalk, line, pte, and node2vec,'' in
  \emph{WSDM}, 2018.

\bibitem{wang2019neural}
X.~Wang, X.~He, M.~Wang, F.~Feng, and T.-S. Chua, ``Neural graph collaborative
  filtering,'' in \emph{Proceedings of the 42nd international ACM SIGIR
  conference on Research and development in Information Retrieval}, 2019, pp.
  165--174.

\bibitem{LiangTNNLS}
L.~Li, S.~Wang, X.~Liu, E.~Zhu, L.~Shen, K.~Li, and K.~Li, ``Local
  sample-weighted multiple kernel clustering with consensus discriminative
  graph,'' \emph{IEEE Transactions on Neural Networks and Learning Systems},
  2022.

\bibitem{ou2016asymmetric}
M.~Ou, P.~Cui, J.~Pei, Z.~Zhang, and W.~Zhu, ``Asymmetric transitivity
  preserving graph embedding,'' in \emph{SIGKDD}, 2016.

\bibitem{li2021learning}
Z.~Li, H.~Liu, Z.~Zhang, T.~Liu, and N.~N. Xiong, ``Learning knowledge graph
  embedding with heterogeneous relation attention networks,'' \emph{IEEE
  Transactions on Neural Networks and Learning Systems}, 2021.

\bibitem{zhang2021we}
S.~Zhang, H.~Chen, X.~Ming, L.~Cui, H.~Yin, and G.~Xu, ``Where are we in
  embedding spaces?'' in \emph{SIGKDD}, 2021.

\bibitem{bo2020structural}
D.~Bo, X.~Wang, C.~Shi, M.~Zhu, E.~Lu, and P.~Cui, ``Structural deep clustering
  network,'' in \emph{Proceedings of the web conference 2020}, 2020, pp.
  1400--1410.

\bibitem{huang2022discovering}
C.~Huang, Q.~Zhang, D.~Guo, X.~Zhao, and X.~Wang, ``Discovering association
  rules with graph patterns in temporal networks,'' \emph{Tsinghua Science and
  Technology}, 2022.

\bibitem{hawkes1971point}
A.~G. Hawkes, ``Point spectra of some mutually exciting point processes,''
  \emph{Journal of the Royal Statistical Society: Series B (Methodological)},
  1971.

\bibitem{wu2022gcrec}
B.~Wu, X.~He, Q.~Zhang, M.~Wang, and Y.~Ye, ``Gcrec: Graph-augmented capsule
  network for next-item recommendation,'' \emph{IEEE Transactions on Neural
  Networks and Learning Systems}, 2022.

\bibitem{hu2023scdfc}
D.~Hu, K.~Liang, S.~Zhou, W.~Tu, M.~Liu, and X.~Liu, ``scdfc: A deep fusion
  clustering method for single-cell rna-seq data,'' \emph{Briefings in
  Bioinformatics}, 2023.

\bibitem{liang2023learn}
K.~Liang, L.~Meng, M.~Liu, Y.~Liu, W.~Tu, S.~Wang, S.~Zhou, and X.~Liu, ``Learn
  from relational correlations and periodic events for temporal knowledge graph
  reasoning,'' in \emph{Proceedings of the 46th International ACM SIGIR
  Conference on Research and Development in Information Retrieval}, 2023.

\bibitem{ji2021survey}
S.~Ji, S.~Pan, E.~Cambria, P.~Marttinen, and S.~Y. Philip, ``A survey on
  knowledge graphs: Representation, acquisition, and applications,'' \emph{IEEE
  transactions on neural networks and learning systems}, 2021.

\bibitem{meng2023sarf}
L.~Meng, K.~Liang, B.~Xiao, S.~Zhou, Y.~Liu, M.~Liu, X.~Yang, and X.~Liu,
  ``Sarf: Aliasing relation assisted self-supervised learning for few-shot
  relation reasoning,'' \emph{arXiv preprint arXiv:2304.10297}, 2023.

\bibitem{gao2023spatiotemporal}
X.~Gao, X.~Jiang, D.~Zhuang, H.~Chen, S.~Wang, and J.~Haworth, ``Spatiotemporal
  graph neural networks with uncertainty quantification for traffic incident
  risk prediction,'' \emph{arXiv preprint arXiv:2309.05072}, 2023.

\bibitem{CCGC}
X.~Yang, Y.~Liu, S.~Zhou, S.~Wang, W.~Tu, Q.~Zheng, X.~Liu, L.~Fang, and
  E.~Zhu, ``Cluster-guided contrastive graph clustering network,'' in
  \emph{Proceedings of the AAAI conference on artificial intelligence}, 2023.

\bibitem{liang2022reasoning}
K.~Liang, L.~Meng, M.~Liu, Y.~Liu, W.~Tu, S.~Wang, S.~Zhou, X.~Liu, and F.~Sun,
  ``Reasoning over different types of knowledge graphs: Static, temporal and
  multi-modal,'' \emph{arXiv preprint arXiv:2212.05767}, 2022.

\bibitem{liang2023structure}
K.~Liang, S.~Zhou, Y.~Liu, L.~Meng, M.~Liu, and X.~Liu, ``Structure guided
  multi-modal pre-trained transformer for knowledge graph reasoning,''
  \emph{arXiv preprint arXiv:2307.03591}, 2023.

\bibitem{DMG_ICML}
Y.~Mo, Y.~Lei, J.~Shen, X.~Shi, H.~T. Shen, and X.~Zhu, ``Disentangled
  multiplex graph representation learning,'' in \emph{International Conference
  on Machine Learning}.\hskip 1em plus 0.5em minus 0.4em\relax PMLR, 2023, pp.
  24\,983--25\,005.

\bibitem{zhou2023multi}
J.~Zhou, J.~Sun, W.~Zhang, and Z.~Lin, ``Multi-view underwater image
  enhancement method via embedded fusion mechanism,'' \emph{Engineering
  Applications of Artificial Intelligence}, 2023.

\bibitem{Liangke_SymCLKG_TKDE}
K.~Liang, Y.~Liu, S.~Zhou, W.~Tu, Y.~Wen, X.~Yang, X.~Dong, and X.~Liu,
  ``Knowledge graph contrastive learning based on relation-symmetrical
  structure,'' \emph{IEEE Transactions on Knowledge and Data Engineering},
  2023.

\bibitem{mo2023multiplex}
Y.~Mo, Y.~Chen, Y.~Lei, L.~Peng, X.~Shi, C.~Yuan, and X.~Zhu, ``Multiplex graph
  representation learning via dual correlation reduction,'' \emph{IEEE
  Transactions on Knowledge and Data Engineering}, 2023.

\bibitem{yu2023g}
H.~Yu, C.~Ma, M.~Liu, T.~Du, M.~Ding, T.~Xiang, S.~Ji, and X.~Liu, ``G2uardfl:
  Safeguarding federated learning against backdoor attacks through attributed
  client graph clustering,'' \emph{arXiv preprint}, 2023.

\bibitem{wang2019heterogeneous}
X.~Wang, H.~Ji, C.~Shi, B.~Wang, Y.~Ye, P.~Cui, and P.~S. Yu, ``Heterogeneous
  graph attention network,'' in \emph{The world wide web conference}, 2019, pp.
  2022--2032.

\bibitem{wu2023hypergraph}
H.~Wu, Y.~Yan, and M.~K.-P. Ng, ``Hypergraph collaborative network on vertices
  and hyperedges,'' \emph{IEEE Transactions on Pattern Analysis and Machine
  Intelligence}, 2023.

\bibitem{wu2023simplicial}
H.~Wu, A.~Yip, J.~Long, J.~Zhang, and M.~K. Ng, ``Simplicial complex neural
  networks,'' \emph{IEEE Transactions on Pattern Analysis and Machine
  Intelligence}, 2023.

\bibitem{fang2022scalable}
Y.~Fang, X.~Zhao, P.~Huang, W.~Xiao, and M.~de~Rijke, ``Scalable representation
  learning for dynamic heterogeneous information networks via metagraphs,''
  \emph{ACM Transactions on Information Systems (TOIS)}, 2022.

\bibitem{gan2022multigraph}
J.~Gan, R.~Hu, Y.~Mo, Z.~Kang, L.~Peng, Y.~Zhu, and X.~Zhu, ``Multigraph fusion
  for dynamic graph convolutional network,'' \emph{IEEE Transactions on Neural
  Networks and Learning Systems}, 2022.

\bibitem{song2022graph}
Z.~Song, X.~Yang, Z.~Xu, and I.~King, ``Graph-based semi-supervised learning: A
  comprehensive review,'' \emph{IEEE Transactions on Neural Networks and
  Learning Systems}, 2022.

\bibitem{Wan_Liu_Liu_Wang_Wen_Liang_Zhu_Liu_Zhou_2023}
X.~Wan, X.~Liu, J.~Liu, S.~Wang, Y.~Wen, W.~Liang, E.~Zhu, Z.~Liu, and L.~Zhou,
  ``Auto-weighted multi-view clustering for large-scale data,''
  \emph{Proceedings of the AAAI Conference on Artificial Intelligence}, 2023.

\bibitem{lin2022prototypical}
S.~Lin, C.~Liu, P.~Zhou, Z.-Y. Hu, S.~Wang, R.~Zhao, Y.~Zheng, L.~Lin, E.~Xing,
  and X.~Liang, ``Prototypical graph contrastive learning,'' \emph{IEEE
  Transactions on Neural Networks and Learning Systems}, 2022.

\bibitem{CONVERT}
X.~Yang, C.~Tan, Y.~Liu, K.~Liang, S.~Wang, S.~Zhou, J.~Xia, S.~Z. Li, X.~Liu,
  and E.~Zhu, ``Convert: Contrastive graph clustering with reliable
  augmentation,'' in \emph{Proceedings of the 31st ACM International Conference
  on Multimedia}, 2023.

\bibitem{perozzi2014deepwalk}
B.~Perozzi, R.~Al-Rfou, and S.~Skiena, ``Deepwalk: Online learning of social
  representations,'' in \emph{SIGKDD}, 2014.

\bibitem{grover2016node2vec}
A.~Grover and J.~Leskovec, ``node2vec: Scalable feature learning for
  networks,'' in \emph{SIGKDD}, 2016.

\bibitem{kipf2016variational}
T.~N. Kipf and M.~Welling, ``Variational graph auto-encoders,'' in
  \emph{NeurIPS}, 2016.

\bibitem{hamilton2017inductive}
W.~Hamilton, Z.~Ying, and J.~Leskovec, ``Inductive representation learning on
  large graphs,'' \emph{NeurIPS}, 2017.

\bibitem{luo2020parameterized}
D.~Luo, W.~Cheng, D.~Xu, W.~Yu, B.~Zong, H.~Chen, and X.~Zhang, ``Parameterized
  explainer for graph neural network,'' \emph{Advances in neural information
  processing systems}, 2020.

\bibitem{xu2021infogcl}
D.~Xu, W.~Cheng, D.~Luo, H.~Chen, and X.~Zhang, ``Infogcl: Information-aware
  graph contrastive learning,'' \emph{Advances in Neural Information Processing
  Systems}, 2021.

\bibitem{zheng2021node}
C.~Zheng, B.~Zong, W.~Cheng, D.~Song, J.~Ni, W.~Yu, H.~Chen, and W.~Wang,
  ``Node classification in temporal graphs through stochastic sparsification
  and temporal structural convolution,'' in \emph{ECML PKDD 2020}, 2021.

\bibitem{cui2022dygcn}
Z.~Cui, Z.~Li, S.~Wu, X.~Zhang, Q.~Liu, L.~Wang, and M.~Ai, ``Dygcn: Efficient
  dynamic graph embedding with graph convolutional network,'' \emph{IEEE
  Transactions on Neural Networks and Learning Systems}, 2022.

\bibitem{TGC_ML}
M.~Liu, Y.~Liu, K.~Liang, S.~Wang, S.~Zhou, and X.~Liu, ``Deep temporal graph
  clustering,'' \emph{arXiv preprint arXiv:2305.10738}, 2023.

\bibitem{pareja2020evolvegcn}
A.~Pareja, G.~Domeniconi, J.~Chen, T.~Ma, T.~Suzumura, H.~Kanezashi, T.~Kaler,
  T.~Schardl, and C.~Leiserson, ``Evolvegcn: Evolving graph convolutional
  networks for dynamic graphs,'' in \emph{AAAI}, 2020.

\bibitem{sankar2020dysat}
A.~Sankar, Y.~Wu, L.~Gou, W.~Zhang, and H.~Yang, ``Dysat: Deep neural
  representation learning on dynamic graphs via self-attention networks,'' in
  \emph{WSDM}, 2020, pp. 519--527.

\bibitem{nguyen2018continuous}
G.~H. Nguyen, J.~B. Lee, R.~A. Rossi, N.~K. Ahmed, E.~Koh, and S.~Kim,
  ``Continuous-time dynamic network embeddings,'' in \emph{Companion
  proceedings of the the web conference 2018}, 2018.

\bibitem{zuo2018embedding}
Y.~Zuo, G.~Liu, H.~Lin, J.~Guo, X.~Hu, and J.~Wu, ``Embedding temporal network
  via neighborhood formation,'' in \emph{SIGKDD}, 2018.

\bibitem{lu2019temporal}
Y.~Lu, X.~Wang, C.~Shi, P.~S. Yu, and Y.~Ye, ``Temporal network embedding with
  micro-and macro-dynamics,'' in \emph{CIKM}, 2019.

\bibitem{xu2019spatio}
D.~Xu, W.~Cheng, D.~Luo, X.~Liu, and X.~Zhang, ``Spatio-temporal attentive rnn
  for node classification in temporal attributed graphs.'' in \emph{IJCAI},
  2019, pp. 3947--3953.

\bibitem{xu2019adaptive}
D.~Xu, W.~Cheng, D.~Luo, Y.~Gu, X.~Liu, J.~Ni, B.~Zong, H.~Chen, and X.~Zhang,
  ``Adaptive neural network for node classification in dynamic networks,'' in
  \emph{2019 IEEE International Conference on Data Mining (ICDM)}, 2019.

\bibitem{xu2020inductive}
D.~Xu, C.~Ruan, E.~Korpeoglu, S.~Kumar, and K.~Achan, ``Inductive
  representation learning on temporal graphs,'' in \emph{ICLR}, 2020.

\bibitem{liu2021inductive}
M.~Liu and Y.~Liu, ``Inductive representation learning in temporal networks via
  mining neighborhood and community influences,'' in \emph{SIGIR}, 2021.

\bibitem{xu2021transformer}
D.~Xu, J.~Liang, W.~Cheng, H.~Wei, H.~Chen, and X.~Zhang, ``Transformer-style
  relational reasoning with dynamic memory updating for temporal network
  modeling,'' in \emph{Proceedings of the AAAI Conference on Artificial
  Intelligence}, 2021.

\bibitem{wen2022trend}
Z.~Wen and Y.~Fang, ``Trend: Temporal event and node dynamics for graph
  representation learning,'' in \emph{Proceedings of the ACM Web Conference
  2022}, 2022.

\bibitem{xu2022dyng2g}
M.~Xu, A.~V. Singh, and G.~E. Karniadakis, ``Dyng2g: An efficient stochastic
  graph embedding method for temporal graphs,'' \emph{IEEE Transactions on
  Neural Networks and Learning Systems}, 2022.

\bibitem{deng2022graph}
L.~Deng, D.~Lian, Z.~Huang, and E.~Chen, ``Graph convolutional adversarial
  networks for spatiotemporal anomaly detection,'' \emph{IEEE Transactions on
  Neural Networks and Learning Systems}, 2022.

\bibitem{duan2023dynamic}
P.~Duan, C.~Zhou, and Y.~Liu, ``Dynamic graph representation learning via
  coupling-process model,'' \emph{IEEE Transactions on Neural Networks and
  Learning Systems}, 2023.

\bibitem{liu2023tmac}
M.~Liu, K.~Liang, D.~Hu, H.~Yu, Y.~Liu, L.~Meng, W.~Tu, S.~Zhou, and X.~Liu,
  ``Tmac: Temporal multi-modal graph learning for acoustic event
  classification,'' in \emph{Proceedings of the 31st ACM International
  Conference on Multimedia}, 2023, pp. 3365--3374.

\bibitem{wang2023dyexplainer}
T.~Wang, D.~Luo, W.~Cheng, H.~Chen, and X.~Zhang, ``Dyexplainer: Explainable
  dynamic graph neural networks,'' \emph{arXiv preprint arXiv:2310.16375},
  2023.

\bibitem{hu2020graph}
L.~Hu, C.~Li, C.~Shi, C.~Yang, and C.~Shao, ``Graph neural news recommendation
  with long-term and short-term interest modeling,'' \emph{Information
  Processing and Management}, 2020.

\bibitem{liu2022embtemporal}
M.~Liu, Z.-W. Quan, J.-M. Wu, Y.~Liu, and M.~Han, ``Embedding temporal networks
  inductively via mining neighborhood and community influences,'' \emph{Applied
  Intelligence}, 2022.

\bibitem{granovetter1978threshold}
M.~Granovetter, ``Threshold models of collective behavior,'' \emph{American
  journal of sociology}, 1978.

\bibitem{li2018infmax}
Y.~Li, J.~Fan, Y.~Wang, and K.-L. Tan, ``Influence maximization on social
  graphs: A survey,'' \emph{IEEE Transactions on Knowledge and Data
  Engineering}, 2018.

\bibitem{tian2011new}
J.-T. Tian, Y.-T. Wang, and X.-J. Feng, ``A new hybrid algorithm for influence
  maximization in social networks,'' \emph{Jisuanji Xuebao(Chinese Journal of
  Computers)}, 2011.

\bibitem{perez2018film}
E.~Perez, F.~Strub, H.~De~Vries, V.~Dumoulin, and A.~Courville, ``Film: Visual
  reasoning with a general conditioning layer,'' in \emph{Proceedings of the
  AAAI Conference on Artificial Intelligence}, 2018.

\bibitem{chen2018pme}
H.~Chen, H.~Yin, W.~Wang, H.~Wang, Q.~V.~H. Nguyen, and X.~Li, ``Pme: projected
  metric embedding on heterogeneous networks for link prediction,'' in
  \emph{SIGKDD}, 2018.

\bibitem{mikolov2013distributed}
T.~Mikolov, I.~Sutskever, K.~Chen, G.~S. Corrado, and J.~Dean, ``Distributed
  representations of words and phrases and their compositionality,''
  \emph{NeurIPS}, vol.~26, 2013.

\bibitem{kumar2019predicting}
S.~Kumar, X.~Zhang, and J.~Leskovec, ``Predicting dynamic embedding trajectory
  in temporal interaction networks,'' in \emph{SIGKDD}, 2019.

\bibitem{panzarasa2009patterns}
P.~Panzarasa, T.~Opsahl, and K.~M. Carley, ``Patterns and dynamics of users'
  behavior and interaction: Network analysis of an online community,''
  \emph{Journal of the American Society for Information Science and
  Technology}, 2009.

\bibitem{leskovec2005graphs}
J.~Leskovec, J.~Kleinberg, and C.~Faloutsos, ``Graphs over time: densification
  laws, shrinking diameters and possible explanations,'' in \emph{SIGKDD},
  2005.

\bibitem{kumar2018rev2}
S.~Kumar, B.~Hooi, D.~Makhija, M.~Kumar, C.~Faloutsos, and V.~Subrahmanian,
  ``Rev2: Fraudulent user prediction in rating platforms,'' in \emph{WSDM},
  2018.

\bibitem{ni2019justifying}
J.~Ni, J.~Li, and J.~McAuley, ``Justifying recommendations using
  distantly-labeled reviews and fine-grained aspects,'' in \emph{EMNLP-IJCNLP},
  2019.

\bibitem{kingma2014adam}
D.~P. Kingma and J.~Ba, ``Adam: A method for stochastic optimization,'' in
  \emph{ICLR}, 2014.

\end{thebibliography}

\vfill

\end{document}